%% file: main.tex
\documentclass{article}

\usepackage{booktabs} 

\newcommand\ie{\textit{i.e.}}
\newcommand\eg{\textit{e.g.}}
\newcommand\cf{\textit{cf.}}
\usepackage[accepted]{icml2024}
\usepackage{enumitem}
\setlist{leftmargin=15pt} 
\setlist[itemize]{noitemsep, topsep=0pt}

\usepackage{amsmath}
\usepackage{amssymb}
\usepackage{mathtools}
\usepackage{amsthm}
\definecolor{cvprblue}{rgb}{0.21,0.49,0.74}
\usepackage[pagebackref,breaklinks,colorlinks,allcolors=cvprblue]{hyperref}
\usepackage[capitalize,noabbrev]{cleveref}

\usepackage{caption}    
\usepackage{subcaption} 
\usepackage{array}
\usepackage{bbm}
\input{preamble}

\icmltitlerunning{CUPS: Improving Human Pose-Shape Estimators with Conformalized Deep Uncertainty}

\begin{document}

\twocolumn[
\icmltitle{CUPS: Improving Human Pose-Shape Estimators \\ with Conformalized Deep Uncertainty}



\icmlsetsymbol{equal}{*}

\begin{icmlauthorlist}
\icmlauthor{Harry Zhang}{sch}
\icmlauthor{Luca Carlone}{sch}
\end{icmlauthorlist}

\icmlaffiliation{sch}{Massachusetts Institute of Technology}

\vskip 0.3in
]



\input{sec/0_abstract}    
\input{sec/1_intro}
\input{sec/2_rw}
\input{sec/3_problem-form}
\input{sec/4_method}
\input{sec/5_experiments}

\input{sec/6_conclusion}

\bibliographystyle{icml2024}
\bibliography{arxiv}

\clearpage
\onecolumn

\begin{center}
    {\Large \bf Supplementary Material}
\end{center}
\setcounter{section}{0}
\input{sec/appendix}

\end{document}

%% file: preamble.tex
\usepackage{thmtools,thm-restate, multirow, multicol,xspace}
\usepackage{color,soul}

\newcommand{\dtv}[2]{D_{\text{TV}}\left(#1 \parallel #2\right)}
\newcommand{\dtvv}[2]{D^2_{\text{TV}}\left(#1 \parallel #2\right)}
\newcommand{\bs}[1]{\boldsymbol{#1}}
\let\proof\relax 
\let\endproof\relax

\newtheorem{theorem}{Theorem}
\newtheorem{lemma}[theorem]{Lemma}

\newtheorem{remark}{Remark}

\newtheorem{definition}{Definition}

\newtheorem{assumption}{Assumption}


%% file: sec/0_abstract.tex

\begin{abstract}
We introduce CUPS, a novel method for learning sequence-to-sequence 3D human shapes and poses from RGB videos with uncertainty quantification. To improve on top of prior work, we develop a method to {generate and score multiple hypotheses during training}, effectively integrating uncertainty quantification into the learning process. This process results in a deep uncertainty function that is trained end-to-end with the 3D pose estimator. Post-training, the learned deep uncertainty model is used as the conformity score, which can be used to calibrate a conformal predictor in order to {assess} the quality of the output prediction. Since the data in human pose-shape learning is not fully exchangeable, we also {present} two practical bounds for the coverage gap in conformal prediction, developing theoretical backing for the uncertainty bound of our model. Our results indicate that
by taking advantage of deep uncertainty with conformal prediction, our method achieves state-of-the-art performance across various
metrics and datasets while inheriting the probabilistic guarantees of conformal prediction. Interactive 3D visualization, code, and data will be available at \href{https://sites.google.com/view/champpp}{\textbf{this website}}.
\end{abstract}

%% file: sec/1_intro.tex

\section{Introduction}
\label{sec:intro}
Recovering a sequence of human meshes (\ie, shapes and poses) from a monocular video is a fundamental challenge with wide-ranging applications in computer vision, robotics, AR/VR, and computer graphics. Such technology has the potential to minimize reliance on traditional motion capture systems or labor-intensive 3D annotations, facilitating the generation of human motion templates for tasks such as animating 3D avatars. The emergence of parametrized human models, such as SMPL \cite{Kanazawa18cvpr-smplify}, which represent human body shape and pose with well-defined joint and structure parameters, made it possible for modern deep learning models to efficiently learn to predict human poses and shapes in a systematic way by directly regressing SMPL parameters from video inputs. 
At the same time, {safety-critical} applications, including robotics and autonomous vehicles, demand computer vision algorithms that are able to quantify the 
uncertainty in their estimates and possibly provide performance guarantees~\cite{Yang23cvpr-object}.
\begin{figure}[t]
    \centering
    \includegraphics[width=\linewidth]{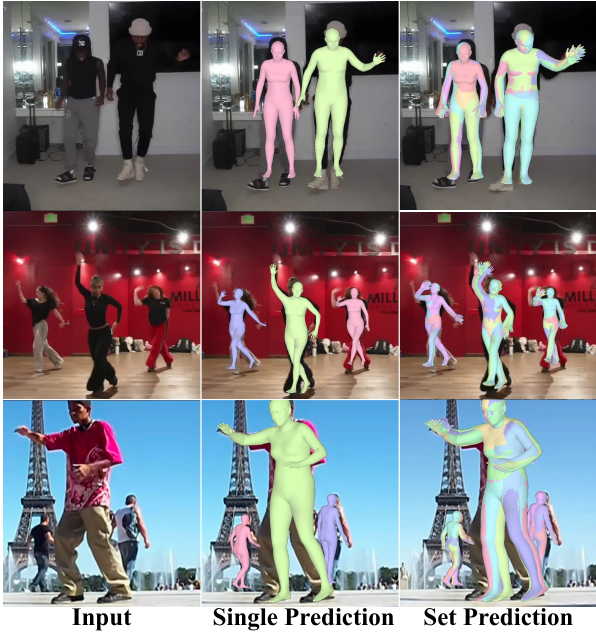}
    \caption{\small CUPS sample results obtained on in-the-wild videos collected from TikTok. Given a sequence of 2D RGB frames, CUPS reconstructs a sequence of 3D human meshes, and then a conformal predictor calibrated using a deep uncertainty function ---trained end-to-end with the human pose-shape estimator--- quantifies the uncertainty of the output SMPL parameters.}
    \label{fig:teaser}
    \vspace{-10pt}
\end{figure}

 Few existing works in human reconstruction have explored the direction of uncertainty-aware human pose and shape prediction due to two challenges. First, it is difficult to ensure the predicted pose and shape are close to the groundtruth under out-of-distribution data or heavy occlusions. 
 Second, an efficient human shape and pose prediction model takes video frames as input, where the data is not fully exchangeable. Such non-exchangeability makes it difficult for uncertainty quantification methods such as conformal prediction \cite{Angelopoulos21gentle, Shafer08jmlr-tutorial} to provide a formal statistical error bound between the estimation and the groundtruth. While a more recent line of work focuses on learning multiple outputs or learning variances as uncertainty \cite{Zhang24arxiv-CHAMP}, none has addressed the problem of providing a reliable statistical error bound when the data is not exchangeable. For practical uses in safety-critical scenarios, one should be able to tell when to \textit{trust} the human reconstruction model.

To counter the aforementioned challenges, we first take inspiration from an important tool in statistical learning, conformal prediction (CP) \citep{Shafer08jmlr-tutorial, Angelopoulos21gentle}, which uses a \textit{post-training} calibration step to guarantee a user-specified coverage. Assume that we want to predict an output Y (\eg, the true pose and shape of a human) from inputs X (\eg, a sequence of frames). By allowing to predict \emph{confidence sets} $C(X)$ (\eg, a set of human reconstructions), CP guarantees the true value $Y$ to be included in $C(X)$ with confidence level $\alpha$, \ie, $P(Y \in C(X)) \geq 1 - \alpha$, given a set of calibration examples $(X_i, Y_i) \in I_\text{cal}$ that are \textit{exchangeable} with the test distribution. There are typically two steps involved in CP. In the calibration step, the conformity scores of the examples in the calibration set are ranked to determine a cut-off threshold $\mathbb{Q}_{1-\alpha}$, via quantile computation. 
In the prediction step, the conformity score measures the conformity between the output and the unknown ground-truth value, which is used ---in conjunction with the threshold $\mathbb{Q}_{1-\alpha}$--- to construct the confidence sets $C(X)$. By construction, the set $C(X)$ provides a quantification of uncertainty: each prediction in $C(X)$ has a conformity score above $(1-\alpha)$th quantile and hence is a plausible estimate for Y.

While CP is a flexible tool that can be applied to any machine learning model, it assumes the data to be \emph{exchangeable}, \ie, that the 
dataset distribution remains invariant under permutation.
 Such an assumption breaks in many real-world applications. For example, if the dataset comes from video frames, the constructed dataset is obviously not exchangeable since permuting sequences of frames changes the underlying distribution. While there have been techniques that aim to increase the exchangeability of video data by taking long video sequences \cite{Zhang24arxiv-CHAMP} or observations from evenly-spaced cameras \cite{Yang23cvpr-object}, the theoretical guarantee of CP cannot be fully justified if the data is not exchangeable. 
To cope with the lack of exchangeability, we leverage a recent extension by \citet{Barber23aos-conformal} that allows dealing with datasets where the exchengeability assumption no longer holds. The key idea in \citet{Barber23aos-conformal} is to use weighted quantiles to tackle data distribution shifts.  

To bring CP into human reconstruction,
 we design a methodology that learns a \textit{deep uncertainty score} of human reconstruction output in an end-to-end manner by predicting multiple hypotheses of human shapes and poses during training. {The deep uncertainty value is then used in the calibration step 
 by incorporating the theoretical toolkit provided in {\cite{Barber23aos-conformal}}, retaining statistical guarantees even when exchangeability is violated.
  Our results indicate that taking advantage of deep uncertainty with conformal prediction, our method achieves state-of-the-art performance across various metrics and datasets. Using the probabilistic guarantee of correctness inherited from CP, we also provide theoretical lower bounds of performance for human mesh reconstruction when the data is not exchangeable.} 
 The result is 
  \textbf{CUPS}, a \textbf{C}onformalized \textbf{U}ncertainty-aware human \textbf{P}ose-\textbf{S}hape estimator. To summarize, our contributions include:
\begin{itemize}
    \item An uncertainty-aware 3D human shape-pose estimator from 2D RGB videos (\cref{sec:human-model} \& \cref{sec:duf}).
    \item A novel method to conformalize 3D human estimates during training by learning a score function to rank the uncertainty of the proposed estimates (\cref{sec:duf}).
    \item A novel uncertainty quantifier for human reconstruction outputs using non-exchangeable conformal prediction and deep uncertainty function (\cref{sec:conformalhuman}).
    \item A theoretical analysis of the uncertainty when the data is not fully exchangeable and two practical lower bounds {---one completely new and one adapted from~\cite{Barber23aos-conformal}---} for the coverage performance (\cref{sec:conformalhuman}).
    \item Quantitative and qualitative results that demonstrate the state-of-the-art results of our method on a variety of real-world datasets (\cref{sec:exp}).
\end{itemize}

%% file: sec/2_rw.tex

\section{Related Work}
\label{sec:rw}
\textbf{3D Human Shape and Pose Estimation.} 
End-to-end approaches for human pose estimation include
\citep{Pavlakos17cvpr-coarse, Sun18eccv-integral}. With the maturity of 2D human keypoints detection \citep{Ho22jmlr-cascaded, Ma22aaai-remote}, more robust approaches focus on lifting 2D keypoints to 3D, resulting in better performance \citep{Xu21cvpr-graph, Ma21cvpr-context, Ci19iccv-optimizing}. In this scheme, deterministic methods learn to predict one single 3D output from the 2D input \citep{Zhan22cvpr-ray3d, Zhang22cvpr-mixste}.
In many applications, it is desirable to also recover the \emph{shape} of humans beyond a skeleton of keypoints.
\cite{Loper23sg-smpl, Kanazawa18cvpr-smplify} propose SMPL, a universal parametrization for human pose and shape. 
MEVA \cite{Luo20accv-meva} utilizes a VAE to encode the motion sequence and generate coarse human mesh sequences which are then refined via a residual correction. VIBE \cite{Kocabas20cvpr-vibe}, TCMR \cite{Choi21cvpr-tcmr} and MPS-Net \cite{Wei22cvpr-mps} encode representations of three different input lengths and 
then learns the mid-frame of the sequence with either a recurrent network or an attention module. GLoT \cite{Shen23cvpr-glot} is a new model that decouples long-term and short-term correlations. We incorporate human mesh with uncertainty learning during training, forcing the network to output higher-quality meshes.

\textbf{Uncertainty in 3D Human Reconstruction.} Due to uncertainty such as occlusion in RGB inputs, deep generative models have been used in modeling conditional distributions for such problems. Mixed-density network \citep{Li19cvpr-generating}, VAE \citep{Sharma19iccv-monocular}, normalizing flows \citep{Wehrbein21iccv-probabilistic}, GAN \citep{Li19cvpr-generating}, and Diffusion models  \citep{Holmquist23iccv-diffpose, Shan23cvpr-diffusion} have all been applied to modeling such conditional distribution. 
\citet{Dwivedi243dv-poco} learn an explicit confidence value for occlusions. Motion-based methods such as \cite{Zhang24cvpr-rohm, Rempe21cvpr-humor} use physical contacts and trajectory consistency to make the uncertain estimates more robust. \citet{Zhang23cvpr-body} use explicit anatomy constraints to improve model performance. Lastly, \citet{Biggs20neurips-3dmb} generate a fixed number of hypotheses and learn to choose the best one.
We learn an uncertainty score by augmenting training outputs and use the uncertainty score to probabilistically certify the outputs.

\textbf{Conformal Prediction.} CP is a powerful and flexible distribution-free uncertainty quantification technique that can be applied to any machine learning model \citep{Angelopoulos21gentle, Shafer08jmlr-tutorial} under the assumption of exchangeable data. Assuming the exchangeability of the calibration data, CP has desirable coverage guarantees. Thus, it has been applied to many fields such as robotics \citep{Sun24neurips-conformal}, pose estimation \citep{Yang23cvpr-object}, and image regression \citep{Angelopoulos22icml-image}. More sophisticated CP paradigms have been also proposed to tackle distribution shift and online learning problems \citep{Angelopoulos24neurips-conformal}. More recently, theoretical grounding for conformal prediction beyond exchangeability assumption \cite{Barber23aos-conformal} has been proposed, which provides analysis tools for non-fully-exchangeable datasets such as videos in ML problems.

%% file: sec/3_problem-form.tex

\section{Problem Formulation}
\label{sec:probform}
We are interested in the problem of learning a sequence of 3D human shapes and poses from a sequence of 2D RGB images. Formally, given the input 2D video sequence $\bs X \in \mathbb{R}^{H\times W\times3\times T}$, where $H, W$ are the dimension of each frame and $T$ is the length of the input sequence, our goal is to learn to output human shapes and poses, 
described by SMPL parameters $\bs Y := \{\bs\theta, \bs\beta\}$, where $\bs\theta \in \mathbb{R}^{24\times6\times T}$ and $\bs\beta\in \mathbb{R}^{10\times T}$ model the joint 6D pose and mesh shape, respectively. We wish to learn a human reconstruction function $f_\theta(\bs X)$ that approximates $\bs Y$. We are also interested in learning a deep uncertainty function $S_\theta(\bs X, \bs Y)$, which measures the inherent uncertainty of the human reconstruction function $f_\theta$ when taking as input $\bs X$ and outputting $\bs Y$. Such a learned uncertainty function will be 
used 
as a conformity score in our method, and enables the construction of a prediction set via conformal prediction.

%% file: sec/4_method.tex
\begin{figure*}[h]
    \centering
    \includegraphics[width=\linewidth]{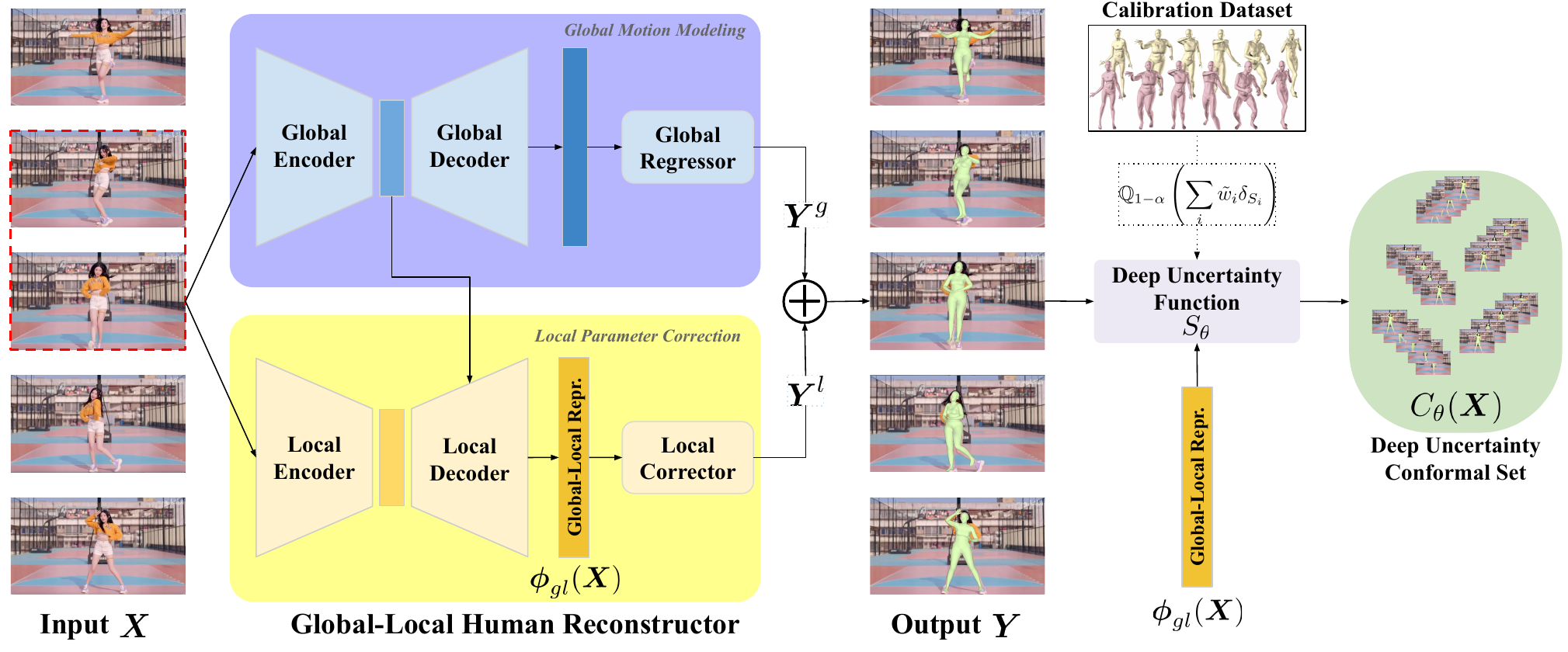}
    \caption{\footnotesize{CUPS Overview. CUPS takes as input a sequence of input RGB video frames. The RGB video frames get encoded and fed into a global-local transformer human reconstruction model to produce SMPL parameters representing the human pose and shape in 3D as well as a decoupled global-local embedding. The output of the human reconstructor is supervised via SMPL loss. While training, we also learn a deep uncertainty function that learns to rank the uncertainty of the produced output sequence. Then after training, this deep uncertainty function is used as the conformity score for constructing a conformal set for conformal prediction.}}
    \label{fig:system}
\end{figure*}

\section{Method}
\label{sec:method}
We use a transformer-based architecture \cite{Shen23cvpr-glot} to predict SMPL parameters from 2D video sequences. We also learn an uncertainty scoring model together with the reconstruction model. At test time,
we use the uncertainty scoring model for conformal prediction. 
 This results in {CUPS}, a {C}onformalized {U}ncertainty-aware human {P}ose-{S}hape estimator. The pipeline of CUPS is shown in \cref{fig:system}.

\subsection{GLoT Human Reconstruction Model}
\label{sec:human-model}
The human reconstruction model in CUPS is based on the Global-to-Local Transformer (GLoT) architecture proposed by \cite{Shen23cvpr-glot}, which robustly leverages information learned with deep networks as well as human prior structures while decoupling the short-term and long-term modeling processes. We summarize the key components proposed in \cite{Shen23cvpr-glot} below.

\textbf{Global Motion Modeling.} First, a pretrained ResNet-50 extracts features from individual frames, resulting in static tokens referred to as $\mathcal{S} = \{\bs s_1, \cdots, \bs s_T\}\in\mathbb{R}^{T\times 2048}$. The global motion modeling step begins by randomly masking a subset of static tokens along the temporal dimension, denoted as $\mathcal{S}^g \in \mathbb{R}^{(1-p)T \times 2048}$, where $p$ represents the mask ratio. The unmasked tokens are then passed through a global encoder. During the global decoder phase, the mean SMPL parameters encoded by an MLP (SMPL tokens) are padded into the masked positions and the entire sequence is fed into the global decoder, which generates a long-term representation. Subsequently, the global motion modeling step applies an iterative regressor \cite{Kanazawa18cvpr-smplify, Kocabas20cvpr-vibe} to obtain the global initial SMPL sequence, denoted by $\bs Y ^g=\{\bs\theta^g, \bs\beta^g\}$.

\textbf{Local Parameter Correction.} A local parameter correction step refines the SMPL parameters outputted by the global modeling step. The local correction step consists of a local transformer and a Hierarchical Spatial Correlation Regressor. 
The local transformer captures short-term local details in neighboring frames: nearby frames' static tokens are selected for short-term modeling, denoted as $\mathcal{S}^l = \left\{ \bs s_t \right\}_{t = \frac{T}{2} - w}^{\frac{T}{2} + w}$ where \( w \) represents the length of the selected neighborhood. When decoding, cross-attention is applied to the query of the mid-frame token and key and value of the global encoder, capturing global human motion consistency and local, fine-grained human mesh structures. 

\textbf{Global-Local Representation. }In the hierarchical spatial correlation regressor step, the model incorporates both the global prediction and a decoupled global-local representation into the regressor.
The model learns joint correlations within the kinematic structure by modeling the local intra-frame human mesh structure, outputting a correction term $\bs Y^l=\{\bs\theta^l, \bs\beta^l\}$, where $\bs \theta^l$ is obtained using an MLP applied on the \textit{decoupled global-local representation} $\bs \phi_{gl}(\bs X)$ and global output $\bs\theta^g$, and $\bs \beta^l$ using $\bs \phi_{gl}(\bs X)$ and $\bs \beta^g$. The human reconstruction model's final output SMPL values are obtained by adding the initial global prediction and the local correction output: $\bs Y := \{\bs\theta, \bs\beta\} = \bs Y^g + \bs Y ^l$.

\textbf{Training Objective. }We follow previous works and apply standard L2 loss to the SMPL parameters and 3D/2D joints location \cite{Kanazawa18cvpr-smplify}. We also follow the velocity loss on 3D/2D joint location proposed in \cite{Shen23cvpr-glot} to learn consistency and capture the long-range dependency. We refer to the combined loss as $\mathcal{L}_G$.

\subsection{Deep Uncertainty Function}
\label{sec:duf}
Conformal predictors are calibrated using a nonconformity score function \cite{Angelopoulos21gentle}. Intuitively, this function measures to which extent a datapoint is unusual relative to a calibration dataset. The most common nonconformity functions are either simple residual terms \cite{Shafer08jmlr-tutorial}, raw logits \cite{Stutz21iclr-learning}, or hand-designed functions \cite{Yang23cvpr-object}. A more recent line of work has demonstrated the benefits of using \textit{learned nonconformity score function} \cite{Zhang24arxiv-CHAMP}, where the score is learned in an end-to-end manner with the machine learning model. In CUPS, we also learn the score end-to-end with the human reconstruction model. Moreover, as the nonconformity score measures how ``unusual'' the datapoint is relative to the calibration set, it provides an inherent uncertainty measure for the model: the lower the nonconformity score, the less uncertain the model is about the datapoint, and vice versa.

Formally, we wish to learn a function $S_\theta(\bs X, \bs Y) \in [0, 1]$ as the nonconformity score, which we refer to as the Deep Uncertainty Function.

\begin{definition}[Deep Uncertainty Function]
\label{def:duf}
The Deep Uncertainty Function takes as input the decoupled global-local representation $\bs \phi_{gl}(\bs X)$ as well as the corrected SMPL parameters $\bs Y$ and outputs a value between 0 and 1 using an MLP:
\begin{equation}
    S_\theta(\bs X, \bs Y) = \sigma\left(\bs\phi^{\mathrm{pred}}\right)\in[0,1]
\end{equation} 
where  $\phi^{\mathrm{pred}} = \mathrm{MLP}(\bs \phi_{gl}(\bs X), \bs\theta, \bs\beta)$ is the output from a multi-layer-perceptron and $\sigma$ is the sigmoid function.
\end{definition}

\textbf{Training Time Ensemble Augmentation.} To better learn the deep uncertainty function, we augment the model output with randomness such that the function learns to rank different samples. We achieve this by utilizing the intrinsic randomness in the human reconstruction model. Note that the global static tokens $\mathcal{S}^g$ (and thus the global-local representation $\bs \phi_{gl}(\bs X)$) is obtained by randomly masking some portion of the input video frames. We augment each training step by randomly masking the video frames for $H_\text{train}$ times, effectively simulating $H_\text{train}$ hypotheses given the same input data. Thus, for each input video sequence $\bs X$, we get multiple hypotheses of SMPL parameters prediction: $\{\bs Y_i\}_{i=1}^{H_\text{train}}$ and global-local representation $\{\bs \phi_{gl}(\bs X)_i\}_{i=1}^{H_\text{train}}$. The proposed samples are used to train  $S_\theta$.

\textbf{Training Objective. }The uncertainty function is implemented as a discriminator-style scoring function that measures the quality of the generated SMPL parameters conditioned on the input sequence, similar to the discriminator loss in \citep{Kocabas20cvpr-vibe}. Thus, a \textit{lower} score function output means the output is \textit{more likely} (more realistic) to be from the ground-truth distribution in the embedding space. Formally, the deep uncertainty function optimizes the following loss:
\begin{equation}
    \mathcal{L}_S = \mathbb{E}\left[S_\theta(\bs X, \bs Y_\text{GT})^2\right]+\mathbb{E}\left[(1-S_\theta(\bs X, \bs Y))^2\right],
\end{equation}
where $\bs Y_\text{GT}$ is the groundtruth SMPL parameters. {Intuitively, this loss makes sure that samples close to the ground truth get higher conformity and vice versa, encouraging the scoring model to discriminate the prediction from the ground truth.}
 Moreover, we also add an adversarial loss that will be back-propagated into the denoiser model, as done in \citep{Kocabas20cvpr-vibe, Zhang24arxiv-CHAMP, Goodfellow20acm-generative}, which encourages the prediction model to output more realistic samples, adversarially confusing the discriminator:
\begin{equation}
     \mathcal{L}_\text{adv} = \mathbb{E}\left[S_\theta(\bs X, \bs Y)^2\right]
\end{equation}
The overall training loss is a sum of SMPL and score loss: $\mathcal{L}_\text{net} = \mathcal{L}_G + \lambda(\mathcal{L}_S+\mathcal{L}_\text{adv})$, where $\lambda$ is a hyperparameter.

\subsection{Conformal Human Reconstruction}
\label{sec:conformalhuman}
Next we use the deep uncertainty function to quantify the uncertainty of the predicted human SMPL output. We first leverage the theoretical toolkit provided by \cite{Barber23aos-conformal}, combining it with CUPS' deep uncertainty function, and then build on top of the coverage guarantee for non-exchangeable conformal prediction in \cite{Barber23aos-conformal} and provide two practical error bounds based on the characteristic of the video dataset and the design choice of the CUPS architecture.

\textbf{SMPL Conformal Calibration. }We introduce the calibration step post-training by using the score in \cref{def:duf}. For any prediction $\bs Y$ from the human reconstruction model and its corresponding input video sequence $\bs X$, $S_\theta(\bs X, \bs Y)$ measures its ``similarity'' to an existing dataset --- the nonconformity score for conformal calibration. When exchangeability holds (formally defined in \cref{app:exc}), the CP calibration is done by choosing a threshold using the $(1-\alpha)$th quantile of the conformity scores calculated on the calibration set. Succinctly, we can define this threshold $\tau$ on the calibration dataset as follows:
\begin{equation}
    \tau = \mathbb{Q}_{1-\alpha} \left(\sum_{i}\delta_{S_\theta(\bs X_i, \bs Y_i)} \right),
\end{equation}
where $\delta_a$ represents the point mass at point $a$ and $\mathbb{Q}$ represents the quantile calculation. However, when the calibration is no longer exchangeable, such a threshold would not yield the desirable coverage guarantee of standard CP. \citet{Barber23aos-conformal} propose to incorporate weighting terms in the CP calibration. Specifically, the new threshold $\tau^*$ is now calculated with prespecified weights $\Tilde{w}$:
\begin{equation}
\label{eq:taustar}
    \tau^* = \mathbb{Q}_{1-\alpha} \left(\sum_{i}\Tilde{w}_i\cdot\delta_{S_\theta(\bs X_i, \bs Y_i)} \right),
\end{equation}
where $\Tilde{w}_i\in[0,1]$ denotes a prespecified weight placed on data point $i$. The values of $\Tilde{w}_i$ is a design choice and it should intuitively be large for data with low non-conformity in the calibration. We discuss some practical design choices for the weights in the sections below.

\textbf{SMPL Conformal Prediction. }Once the threshold value $\tau^*$ is calibrated, we are able to do conformal prediction. Using $\tau^*$ defined in \cref{eq:taustar} and the deep uncertainty function, for a datapoint $\bs X$, we define the Deep Uncertainty Conformal Set (DUCS) as the conformal prediction set.
\begin{restatable}[DUCS]{definition}{ducs}
\label{def:ducs}
The deep uncertainty conformal prediction set is the set of input-output pairs $X,Y$ such that the deep uncertainty value $S_\theta(\bs X, \bs Y)$ is below the calibrated threshold value $\tau^*$
\begin{equation}
    C_\theta(\bs X) = \left\{ \bs Y:\; S_\theta(\bs X, \bs Y)\leq\tau^*\right\}.
\end{equation}
\end{restatable}
For a test video sequence $\bs X$ and predicted SMPL parameters $\bs Y$, it is straightforward to check its set membership in DUCS. {More importantly, as we will show in {\cref{sec:mcdropout}}, we can explicitly make a set prediction by using Monte Carlo Dropout during test time.}

Under the framework of nonexchangeable conformal prediction~\cite{Barber23aos-conformal}, we analyze some theoretical properties of DUCS, which are amenable to our human reconstruction pipeline. 
We first define the tuple $\bs Z_i = (\bs X_i, \bs Y_i)$, which denotes the $i$-th example in our calibration dataset. 
We then construct the following sequence by combining the tuples $\bs Z_i$: $\bs Z = (\bs Z_1, \bs Z_2, \cdots, \bs Z_{n})$ and define
\begin{equation}
\bs Z^i = (\bs Z_1, \cdots, \bs Z_{i-1}, \bs Z_{n}, \cdots, \bs Z_{n-1}, \bs Z_i),
\end{equation}
which represents $\bs Z$ sequence after swapping the last point with the $i$-th calibration datapoint. Now, we define the weights $\Tilde{w}_i$ needed for calibration (\cf~\eqref{eq:taustar}). As mentioned above, the exact weight formulation is a design choice, but to facilitate our analysis, we choose the following design based on the Euclidean distance of learned features.
\begin{restatable}[Feature Distance Weight]{definition}{sdw}
\label{def:weight}
    For our SMPL conformal calibration, the weight is defined based on the feature distance between the predicted SMPL feature and the ground-truth SMPL feature:
    \begin{equation}
    w_i = \exp\left(-\frac{||\phi_i^{\mathrm{pred}} - \phi_i^{\mathrm{GT}}||^2}{\mathcal{T}}\right),
    \end{equation}
    where $\mathcal{T}$ is the temperature hyperparameter, $\phi_i^{\mathrm{pred}} = \mathrm{MLP}(\bs \phi_{gl}(\bs X_i), \bs\theta_i, \bs\beta_i)$ is the predicted embedding in Definition \ref{def:duf} and $\phi_i^{\mathrm{GT}} = \mathrm{MLP}(\bs \phi_{gl}(\bs X_i), \bs\theta_i^{\mathrm{GT}}, \bs\beta_i^{\mathrm{GT}})$ is the ground-truth embedding.
\end{restatable}
Then the quantile weights $\Tilde{w}_i$ is the \textit{normalized} version of $w_i$. The exact normalization technique is again a design choice that we describe below in \cref{eq:normalization}.

We now formally state the coverage guarantee of DUCS using Theorem 2 in \cite{Barber23aos-conformal}. Without assuming exchangeability, DUCS is designed to be robust against distribution shifts.
\begin{restatable}[Nonexchangeable Coverage \protect{\citep[Thm. 2]{Barber23aos-conformal}}]{theorem}{coverage}
\label{thm:coverage}
    Under possibly non-exchangeable dataset distribution, the conformal prediction set defined in Definition \ref{def:ducs} satisfies the following coverage guarantee:
    {\small
    \begin{equation}
    \mathbb{P}\left(\bs Y\in C_\theta(\bs X)\right) \geq 1-\alpha-\sum_{i=1}^{n}\Tilde{w}_i\cdot \dtv{\bs S_\theta(\bs Z)}{\bs S_\theta(\bs Z^i)},
    \end{equation} }
    \noindent
    where $\Tilde{w}_i$ is the normalized weight obtained via Definition \ref{def:weight}, $\dtv{\cdot}{\cdot}$ represents the total variation distance, $\bs S_\theta(\bs Z) = [S_\theta(\bs Z_i)]_{i=1}^{n}$ and similarly for $\bs S_\theta(\bs Z^i)$.
\end{restatable}
We provide a proof in Appendix \ref{app:sec-proof-coverage}, following the outline in~\cite{Barber23aos-conformal}. The extra term $\sum_i^{n}\Tilde{w}_i\cdot \dtv{\bs S_\theta(\bs Z)}{\bs S_\theta(\bs Z^i)}$ is referred to as the \textit{miscoverage gap}.
Next, we {present} two practical bounds for the miscoverage gap by leveraging the structure of the datasets and architectures we use in our mesh estimation problems.

{The first bound is borrowed from \cite{Barber23aos-conformal}}. We leverage the video dataset characteristics, assuming the distribution shift happens periodically. Weights are designed to account for this periodic change.

\begin{restatable}[Miscoverage under Periodic Change  \protect{\citep[4.4]{Barber23aos-conformal}}]{theorem}{gapone}
\label{thm:gap1}
Using $w_i$ in Definition \ref{def:weight}, we define the auxiliary weight ${w}_i'$:
\begin{equation}
    {w}_i' = \rho^{n+1-\pi(w_i)},
\end{equation}
where $\rho$ is a decay hyperparameter and $\pi(w_i)$ maps $w_i$ to its ranked position $ \in [n]$ among all weights. Then the normalized weights are $\Tilde{w}_i = \frac{{w}_i'}{\sum_j {w}_j'}$.
Assuming that the most recent changepoint in the video dataset occurred $k$ time steps ago ---such that $\dtv{\bs Z_i}{\bs Z_n} = 0$ for $i>n-k$ and could be arbitrarily large otherwise--- we have the following bound:
\begin{equation}
\label{eq:normalization}
\sum_{i=1}^{n}\Tilde{w}_i\cdot \dtv{\bs S_\theta(\bs Z)}{\bs S_\theta(\bs Z^i)}\leq \rho^k.
\end{equation}
\end{restatable}
This bound suggests that the miscoverage gap remains small as long as $k$ is large. 

The second bound {is novel} and models the deep uncertainty output on the calibration set $\bs S_\theta(\bs Z)$ using a beta distribution. In practice, this is achieved at test time using Monte Carlo Dropout \cite{Gal16icml-mcdropout}. The bound depends on the beta distributions formed by  $\bs S_\theta(\bs Z)$ and $\bs S_\theta(\bs Z^i)$.

\begin{restatable}[Miscoverage under Beta Distribution]{theorem}{gaptwo}
Assume the deep uncertainty values of the calibration set of size $n$ follow Beta distributions: $\bs S_\theta(\bs Z) \sim \beta(a_1, n-a_1), \quad \bs S_\theta(\bs Z^i) \sim \beta(a_2, n-a_2)$. If we assume that the difference between parameters  $a_1$ and $a_2$  is bounded by $k$, we get the following bound without any assumption on the weights:
\label{thm:gap2}
{\small
\begin{align}
    \sum_{i=1}^{n}\Tilde{w}_i\cdot\dtv{\bs S_\theta(\bs Z)}{\bs S_\theta(\bs Z^i)}&\leq \sqrt{2- 2\left(1-\frac{2k}{{n+k}}\right)^\frac{k}{2}}
\end{align}
}
\end{restatable}
This bound is stronger when $k$ is smaller. Intuitively, if the two distributions formed by swapping are similar then the miscoverage gap will be small. We refer the reader to Appendix \ref{app:secB} for the proof of Theorem \ref{thm:gap1} and \ref{thm:gap2}.

%% file: sec/5_experiments.tex

\section{Experiments}
\label{sec:exp}
We provide a quantitative evaluation of our model against state-of-the-art baselines. We also provide ablation studies and qualitative results to support our  design choices. We follow previous baselines \cite{Shen23cvpr-glot, Kanazawa18cvpr-smplify, Kanazawa19cvpr-learning, Dwivedi243dv-poco, Choi21cvpr-tcmr} and report several intra-frame metrics, including Mean Per Joint Position Error (\textbf{MPJPE}), Procrustes-aligned MPJPE (\textbf{PA-MPJPE}), and Mean Per Vertex Position Error (\textbf{MPVPE}). Following \cite{Shen23cvpr-glot, Choi21cvpr-tcmr}, we also provide a result for the second-order acceleration error (\textbf{Accel}) for the inter-frame smoothness.
\input{figs/table1}

\subsection{Baselines Comparisons}
We follow the same dataset split and setup as done in previous works and evaluated on 3DPW \cite{Von18eccv-3dpw}, Human3.6M \cite{Ionescu13tpami-human3}, and MPII-3DHP \cite{Mehta173dv-monocular}. More details on the construction of training dataset are in \cref{traintestdataset}.
As shown in \cref{tab:sota_video}, our model outperforms state-of-the-art baseline methods. On 3DPW, for example, we outperform GLoT's PA-MPJPE by 1.9mm, MPJPE by 4.5mm, and MPVPE by 4.6mm. While Accel performance was slightly worse off on 3DPW, on the other two datasets, our method surpasses baselines on all metrics. 
Moreover, our method outperforms GLoT by a noticeable margin, {indicating that the deep uncertainty function $S_\theta$ is important in that it forces the human reconstruction to output higher-quality samples during training}. In \cref{sota-img-video}, following previous works \cite{Choi21cvpr-tcmr, Dwivedi243dv-poco, Shen23cvpr-glot}, we measure our method's generalizability to unseen datasets. In this set of experiments, none of the methods uses 3DPW dataset during training. Again, our method outperforms baselines by a noticeable margin.
\input{figs/table2}

\subsection{Ablation Studies}

\textbf{Training Time Ensemble Augmentation.} We augment the training dataset online to better train the deep uncertainty function by leveraging the intrinsic stochasticity (\eg, frame masking) of the human reconstruction model. We compare results as a function of the number of samples ($H$). When $H=1$, there is no augmentation and we are just running the forward pass once. We train several models using a range of $H$ values and evaluate the MPJPE using these models in \cref{fig:ablation1}. {Results suggest that using more proposed samples during training reduces test error overall and the improvement saturates after 30.}

\textbf{Choice of Conformity Score Function. }We compare our proposed deep uncertainty function (\textbf{DUF}) trained using adversarial loss with several different losses in \cref{score-ablation}: a score function augmented with inefficiency loss (\textbf{Ineff.}) \cite{Stutz21iclr-learning} during training and a classifier-style conformity score function (\textbf{Class.}). 
From \cref{score-ablation}, we see that DUF and inefficiency-augmented DUF result in similar performance quantitatively. While the classifier-style loss performs better than without any scoring function quantitatively, the predicted mesh shape on the videos is less realistic. Please refer to \cref{choice} for more details.
\input{figs/ablations}

\textbf{Strength of the Uncertainty Loss.} Finally, we ablate the {hyperparameter} of the training loss for $S_\theta(\bs X, \bs Y)$, $\lambda$, in the overall training objective $\mathcal{L}_\text{net}$. This is an important ablation in that we can find a suitable scale of the loss for the deep uncertainty function to make sure it does not conflict with the pose loss optimization. Results in \cref{score-ablation} suggest that 0.6 is the most efficient strength across all values, as a smaller scale does not train the scoring model sufficiently and a higher scale conflicts with the pose loss.

\subsection{Monte Carlo Dropout}
\label{sec:mcdropout}
\begin{figure}
    \centering
    \includegraphics[width=\linewidth]{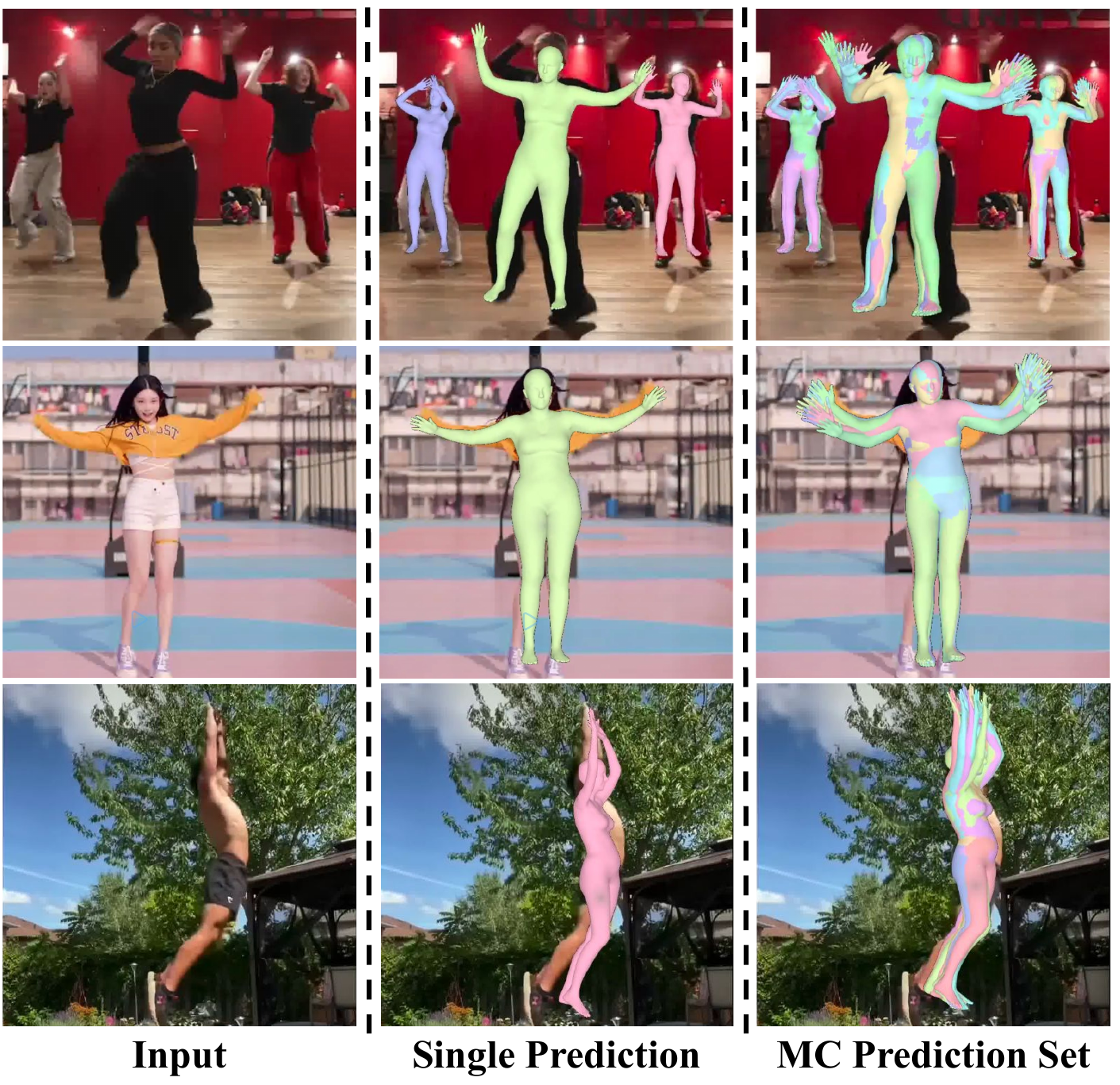}
    \caption{\small In the wild video SMPL predictions with both single hypothesis and multiple hypotheses using MC Dropout.}
    \label{fig:inthewild}
    \vspace{-10pt}
\end{figure}

One interesting byproduct of learning the deep uncertainty function $S_\theta(\bs X, \bs Y)$ is that we can construct the DUCS $C_\theta(\bs X)$ explicitly by sampling the output SMPL parameters multiple times, just like during training time ensemble augmentation. While the model itself is not exactly probabilistic, we can emulate its stochasticity during inference time with Monte Carlo Dropout which lends itself to modeling the model uncertainty in a Bayesian way \cite{Gal16icml-mcdropout}. This procedure effectively enables us to make multi-hypothesis predictions during test time, and by checking the set membership of each hypothesis, we are able to explicitly construct the DUCS $C_\theta(\bs X)$ with minimal changes to the model.
Prediction sets from using MC Dropout are shown in \cref{fig:inthewild}

\subsection{In-the-Wild Videos}
To test the generalizability of our method to in-the-wild videos, we collect videos from YouTube and TikTok. We directly apply the CUPS model trained on the 3DPW dataset to test on in-the-wild videos. We run both regular CUPS and CUPS with MC Dropout for multiple hypotheses results are shown in \cref{fig:inthewild}, where the input videos are collected from TikTok. 
For 3D visualization, please refer to \href{https://sites.google.com/view/champpp}{\textbf{this anonymized website}} to interact with CUPS predictions in 3D.

\subsection{Empirical Coverage}
Here we test the empirical coverage of the deep uncertainty function using the three testing datasets in \cref{tab:sota_video}. Mathematically, we calculate the following value:
\begin{equation}
    \bar{\mathcal{C}} = \frac{1}{|{I}_\text{test}|}\sum_{\bs{Y}^\text{GT}\sim{I}_\text{test}}\mathbbm{1}\left(\boldsymbol{Y}^\text{GT}\in C_\theta(\boldsymbol{X})\right)
\end{equation}
and compare against the desired coverage value 1-$\alpha$.
We use $\alpha = 0.1$ for CP, calibrating with the 90th quantile. Furthermore, we compare the performance with weighted calibration using weights defined in \cref{def:weight} against unweighted calibration as done in regular CP.  From \cref{coverage-ablation}, we see the empirical coverage for weighted CP is around $88\% \pm 3\%$ for all three datasets, which remains close to $1-\alpha$, and in some cases, it exceeds this value. Weighted CP coverage is noticeably higher than unweighted CP, corroborating the results in \cite{Barber23aos-conformal}. The coverage result with weighted CP is encouraging because it illustrates that the miscoverage gap is small in all three datasets.
\begin{table}[h]
    \resizebox{\linewidth}{!}{
      \begin{tabular}{|l|l|l|l|}
        \hline
        & \textbf{3DPW} & \textbf{3DHP} & \textbf{H3.6M} \\ \hline
        \textbf{Weighted CP} & $86.2\pm2.1\%$ & $87.3\pm2.2\%$ & $89.0\pm1.5\%$ \\ \hline
        \textbf{\cref{thm:gap1} Bound} & $\geq83.9\%$ & $\geq84.9\%$ & $\geq85.8\%$ \\ \hline
        \textbf{\cref{thm:gap2} Bound} & $\geq84.0\%$ & $\geq85.3\%$ & $\geq86.8\%$ \\
        \hline
        \textbf{Regular CP} & $81.0\pm3.4\%$ & $83.2\pm2.8\%$ & $85.2\pm2.3\%$ \\
        \hline
      \end{tabular}
    }
    \caption{\small Empirical coverage with weighted vs. regular calibration using the learned deep uncertainty function on three different datasets. Bounds are obtained as described in \cref{app:sec-proof-coverage}.}
    \label{coverage-ablation}
    \vspace{-20pt}
\end{table}

\subsection{Implementation and Training Details}
The global-local human reconstruction model takes as input video sequences of length 16, following \cite{Shen23cvpr-glot, Choi21cvpr-tcmr}. We train with an Adam optimizer with a weight decay parameter of 0.1 and a momentum of 0.9. For the training objective, the adversarial loss weight is 0.6 and is optimized every 100 iterations. Our model is trained using an NVIDIA V100 GPU, where training consumes an amortized GPU memory of 20GB, and CPU memory of 160 GB. We train the model for 100 epochs with an initial learning rate of 5e-5 with a cosine scheduler. During training, the ensemble augmentation step produces 20 samples for the same input datapoint.

%% file: figs/table1.tex
	\begin{table*}[ht]
		\small
		\centering
		
\resizebox{\textwidth}{!}{%
    \begin{tabular}{l|cccc|ccc|ccc|c}
        \toprule[2pt]
        \multirow{2}{*}{\textbf{Method}}& \multicolumn{4}{c|}{\textbf{3DPW \cite{Von18eccv-3dpw}}} & \multicolumn{3}{c|}{\textbf{MPI-INF-3DHP \cite{Wei22cvpr-mps}}} & \multicolumn{3}{c|}{\textbf{Human3.6M \cite{Ionescu13tpami-human3}}} & \multirow{2}{*}{\textbf{\# Frames}} \\
        & PA-MPJPE $\downarrow$ & MPJPE $\downarrow$ & MPVPE $\downarrow$ & Accel $\downarrow$ & PA-MPJPE $\downarrow$ & MPJPE $\downarrow$ & Accel $\downarrow$ & PA-MPJPE $\downarrow$ & MPJPE $\downarrow$ & Accel $\downarrow$ &  \\
        \midrule[1pt]
        \midrule[1pt]
        VIBE~\cite{Kocabas20cvpr-vibe} & 57.6 & 91.9 & - & 25.4 & 68.9 & 103.9 & 27.3 & 53.3 & 78.0 & 27.3 & {16} \\
        MEVA~\cite{Luo20accv-meva} & 54.7 &86.9 &- & 11.6 & 65.4 & 96.4 & 11.1 & 53.2 & 76.0 & 15.3 & 90 \\
        TCMR~\cite{Choi21cvpr-tcmr} & 
        52.7 & 
        86.5 & 102.9& 
        7.1 & 
        63.5& 
        97.3 & 
        8.5 & 
        52.0 & {73.6} & {3.9} & 
        {16} \\
        MPS-Net~\cite{Wei22cvpr-mps} & 
        52.1 & 84.3 & 99.7& 7.4 & 
        62.8 & 96.7 & 9.6 &
        47.4 & 69.4 & {3.6} & {16} \\
        {POCO~\cite{Dwivedi243dv-poco}} &
        {50.5} & {80.5} & {96.5}& {6.7} & 
        {62.1} &
        {93.7} & {8.1} &
        {46.4} & {68.1} & {3.6} & {1}
        \\
        {GLoT~\cite{Shen23cvpr-glot}} &
        {50.6} & {80.7} & {96.3}& {\textbf{6.6}} & 
        {61.5} &
        {93.9} & {7.9} &
        {46.3} & {67.0} & {3.6} & {16}
        \\
    \hline
    {\textbf{CUPS (Ours)}} &
        {\textbf{48.7}} & {\textbf{76.2}} & {\textbf{91.7}}& {{6.9}} & 
        {\textbf{61.3}} &
        {\textbf{92.8}} & {\textbf{7.2}} &
        {\textbf{44.0}} & {\textbf{63.8}} & {\textbf{3.5}} & {16}
        \\
        \bottomrule[2pt]
    \end{tabular}%
}
\caption{\small Multiple errors ($\downarrow$) results on 3DPW, MPI-INF-3DHP, and Human3.6M. All methods use 3DPW training set for training. Comparisons show that CUPS outperforms other baseline methods in the vast majority of metrics.
}
\label{tab:sota_video}
\end{table*}

%% file: figs/table2.tex
\begin{table}[ht]
		\small
		\centering
		
		\resizebox{\linewidth}{!}{%
			\begin{tabular}{ll|cccc}
				\toprule[2pt]
				
				\multicolumn{2}{c}{\multirow{2}{*}{\textbf{Method}}} & \multicolumn{4}{|c}{\textbf{3DPW}} \\
				& & PA-MPJPE $\downarrow$ & MPJPE $\downarrow$ & MPVPE $\downarrow$ & Accel $\downarrow$ \\
				\midrule[1pt]
				\midrule[1pt]
				&HMR~\cite{Kanazawa18cvpr-smplify} & 76.7 & 130.0 & - & 37.4 \\
				&3DMB~\cite{Biggs20neurips-3dmb} & 74.9 & 120.8 & - & - \\
				&SPIN~\cite{Kolotouros19iccv-spin} &  59.2 & 96.9 & 116.4 & 29.8 \\
				&HMMR~\cite{Kanazawa19cvpr-learning} & 72.6 & 116.5 & 139.3 & 15.2 \\
				&VIBE~\cite{Kocabas20cvpr-vibe} & 56.5 & 93.5 & 113.4 & 27.1 \\
				&TCMR~\cite{Choi21cvpr-tcmr} & 55.8 & 95.0 & 111.5 & {7.0} \\ 
				&MPS-Net~\cite{Wei22cvpr-mps} & {54.0} & 91.6 & 109.6 & {7.5} \\
				&POCO \cite{Dwivedi243dv-poco} & {54.7} & {89.3} & {108.4} & {6.8} \\
				&GLoT \cite{Shen23cvpr-glot} & {53.5} & {89.9} & {107.8} & {6.7} \\
    \hline
    &\textbf{CUPS (Ours)} &\textbf{53.0}& \textbf{85.7}& \textbf{103.6}& \textbf{6.6} \\
				
				\bottomrule[2pt]
				
			\end{tabular}%
   }
			\caption{
				\small Multiple errors ($\downarrow$) results on 3DPW. None of the methods use 3DPW for training. CUPS outperforms all baselines.
			}
			\label{sota-img-video}
		\vspace{-20pt}
		\end{table}

%% file: figs/ablations.tex


  
    



\begin{figure*}[h]
\centering

\begin{minipage}{0.32\textwidth}
    \centering
    \includegraphics[width=\linewidth]{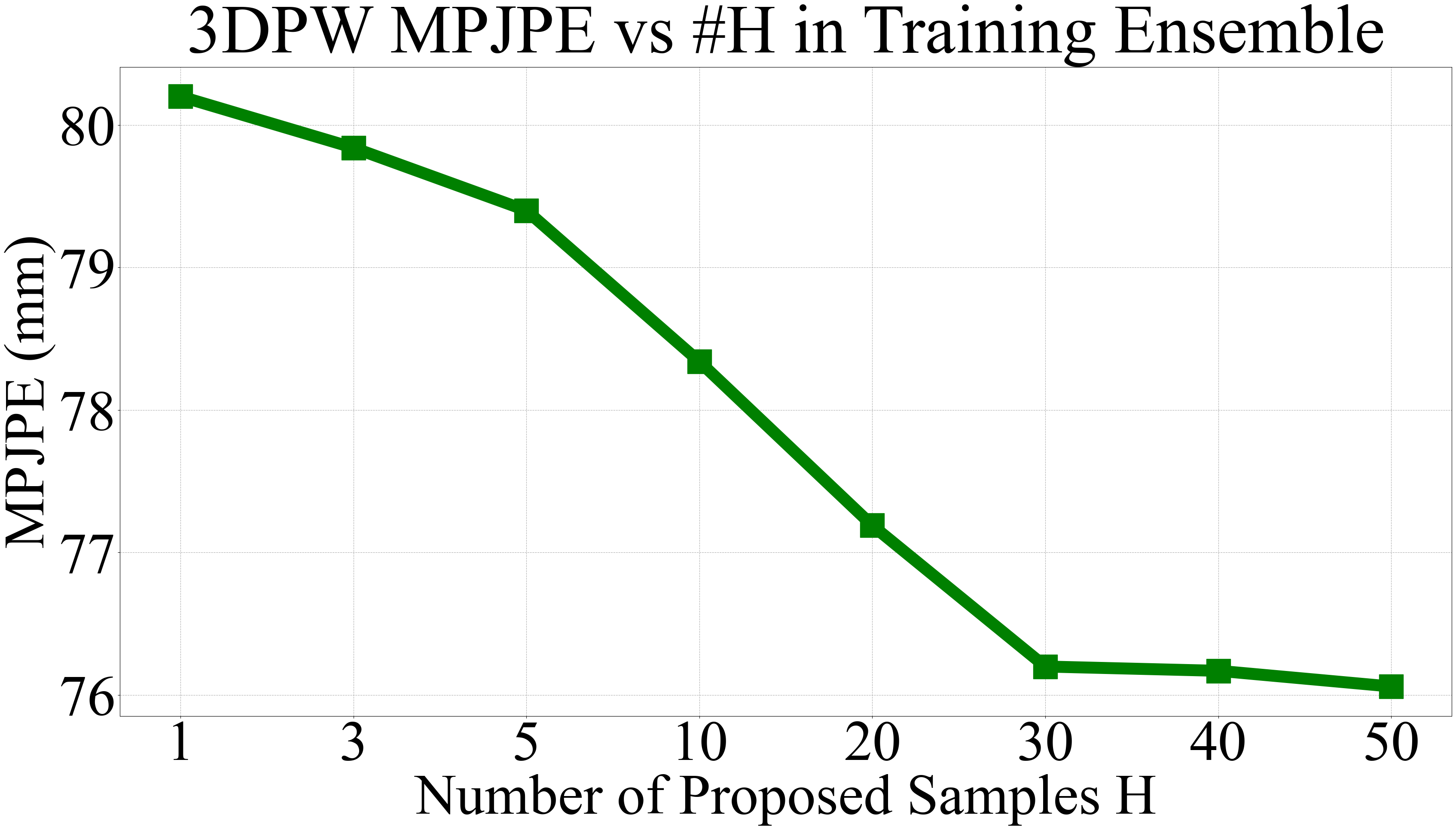}
    \caption{\small Comparison of nr. of samples proposed during training time ensemble.}
    \label{fig:ablation1}
\end{minipage}%
\hfill
\begin{minipage}{0.32\textwidth}
    \centering
    \includegraphics[width=\linewidth]{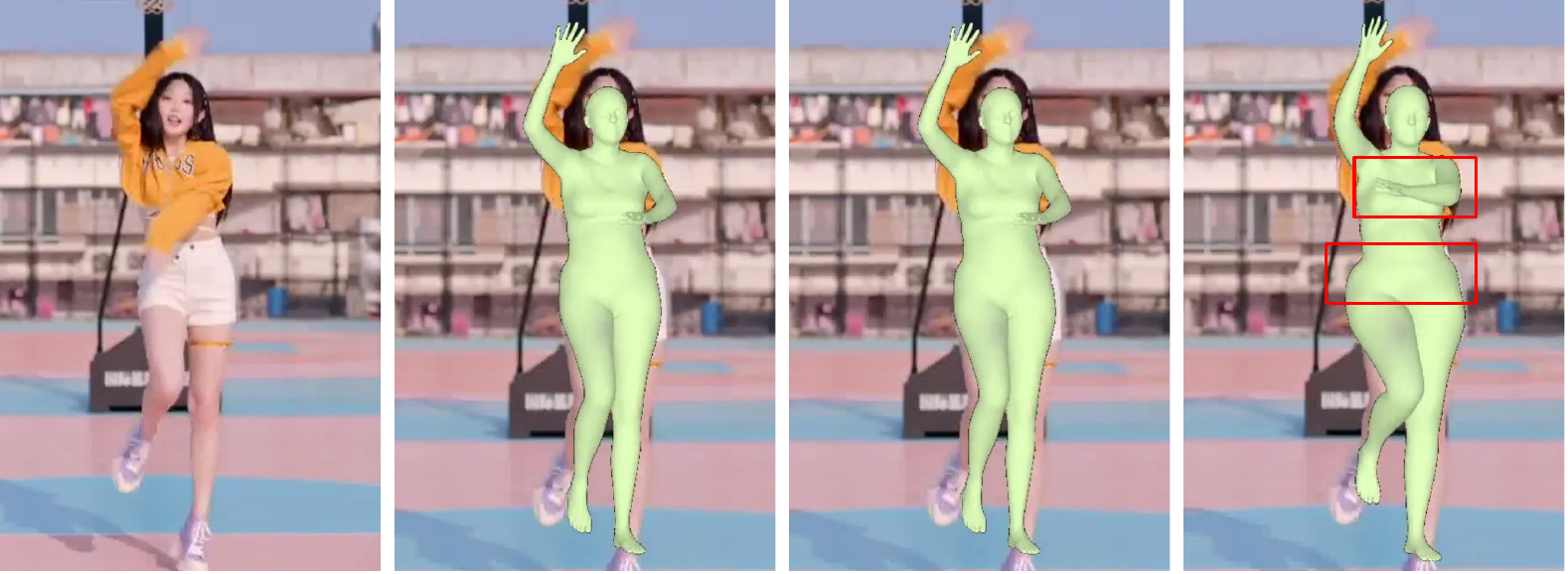} 
    \vspace{5pt}
    \resizebox{\linewidth}{!}{
      \begin{tabular}{|l|l|l|l|}
        \hline
        & \textbf{DUF} & \textbf{Ineff.} & \textbf{Class.} \\ \hline
        \textbf{MPJPE} & 76.2 & 76.2 & 77.4 \\ \hline
      \end{tabular}
    }
    \caption{\small Conformity scores choices on 3DPW (bottom) and internet videos (top).}
    \label{score-ablation}
\end{minipage}%
\hfill
\begin{minipage}{0.32\textwidth}
    \centering
    \includegraphics[width=\linewidth]{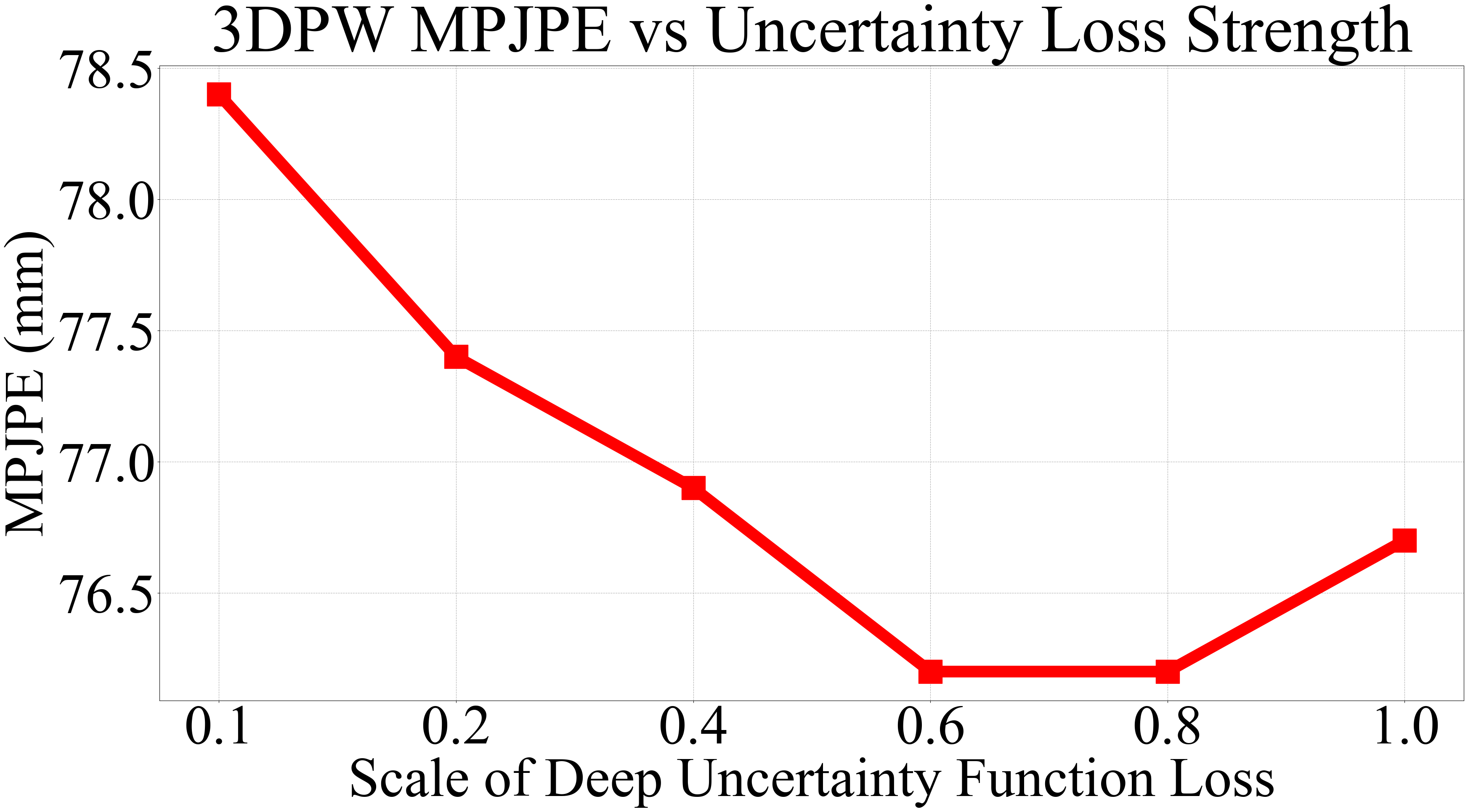}
    \caption{\small Comparison of strength of uncertainty loss in the total training loss.}
    \label{lamb-ablation}
\end{minipage}
\vspace{-10pt}
\end{figure*}

%% file: sec/6_conclusion.tex

\section{Conclusion}
\label{sec:conclusion}
We presented CUPS, an approach for learning sequence-to-sequence 3D human shapes and poses from RGB videos with uncertainty quantification. Our method uses a deep uncertainty function that is trained end-to-end with the 3D pose-shape estimator.
 The deep uncertainty function computes a conformity score, enabling the calibration of a conformal predictor to assess the quality of output predictions at inference time. We {present} two practical bounds for the miscoverage gap in CP, providing theoretical backing for the uncertainty quantification of our model. Our results demonstrate that CUPS achieves state-of-the-art performance across various metrics and datasets, while inheriting the probabilistic guarantees of conformal prediction. We discuss the limitations of CUPS in \cref{app:limitation}.

%% file: sec/appendix.tex
\appendix
\section{Exchangeable Distributions}
\label{app:exc}
First, we define exchangeability in a probabilistic distribution:
\begin{definition}[Exchangeability in Probabilistic Distribution]
    A sequence of random variables 
\( X_1, X_2, \dots, X_n \) is said to have an \textit{exchangeable distribution} 
if the joint distribution of \( X_1, X_2, \dots, X_n \) is invariant under any 
permutation of indices. Formally, for any permutation \( \pi \) of \( \{1, 2, \dots, n\} \),
\[
P(X_1 = x_1, X_2 = x_2, \dots, X_n = x_n) = P(X_{\pi(1)} = x_1, X_{\pi(2)} = x_2, \dots, X_{\pi(n)} = x_n),
\]
for all possible values \( x_1, x_2, \dots, x_n \) of \( X_1, X_2, \dots, X_n \).
\end{definition}
\section{Proof of Coverage Guarantee without Exchangeability}
\label{app:sec-proof-coverage}
We first recall the following definition of Deep Uncertainty Conformal Set:
\ducs*
We further define tuple 
$\bs Z_i = (\bs X_i, \bs Y_i)$ and
\[
\bs Z = (\bs Z_1, \cdots, \bs Z_{n+1}),
\]
as well as 
\[
\bs Z^i = (\bs Z_1, \cdots, \bs Z_{i-1}, \bs Z_{n+1}, \cdots, \bs Z_{n}, \bs Z_i),
\]
which represents $\bs Z$ sequence after swapping the test point with the $i$-th calibration point.
\begin{lemma}[Weight sum upper bound \protect{\citep[Lemma 3]{Harrison12conservative}}] 
\label{lemma:simplebound}
For all $\alpha, w_1, \cdots, w_n\in[0, \infty]$, and all $t_1, \cdots, t_{n+1}\in[-\infty, \infty]$, we have:
\[
\sum_{k=1}^{n+1}w_k\mathbf{1}\left(\sum_{i=1}^{n+1}w_i\mathbf{1}(t_i\geq t_k)\leq \alpha\right)\leq\alpha
\]
\end{lemma}
\emph{Proof.}
\proof
We follow the sketch proof in \cite{Harrison12conservative} and provide detailed proof for interested readers. The tuples $(t_i, w_i)$ can be permuted without affecting the value of the LHS, so we can assume that $t_1\geq\cdots\geq t_n$. This implies that $\sum_{i=1}^{n+1}w_i\mathbf{1}(t_i\geq t_k)$ is increasing in $k$. There exists a $k^*$ defined as follows:
\[
k^*= \sup_k \sum_{i=1}^{n+1}w_i\mathbf{1}(t_i\geq t_k)\leq \alpha
\]
Thus, we have:
\begin{equation}
    \sum_{k=1}^{n+1}w_k\mathbf{1}\left(\sum_{i=1}^{n+1}w_i\mathbf{1}(t_i\geq t_k)\leq \alpha\right)= \sum_{k=1}^{k^*}w_k = \sum_{i=1}^{k^*}w_i\mathbf{1}(t_i\geq t_{k^*})\leq\alpha
\end{equation}
\endproof
Next, recall the nonexchangeable conformal prediction coverage guarantee:
\coverage*

\emph{Proof.}
\proof
We follow the sketch proof in \cite{Barber23aos-conformal} and provide detailed proof for interested readers. We look at the complement of the above probability $\mathbb{P}\left(\bs Y_{n+1}\in C_\theta(\bs X_{n+1})\right)$. For simplicity, we first define $S_i:= S_\theta(\bs X_i, \bs Y_i)$. By definition of DUCS, we have:
\begin{align}
    \begin{split}
        \bs Y_{n+1}\notin C_\theta(\bs X_{n+1})&\Leftrightarrow S_{n+1}>\mathbb{Q}_{1-\alpha} \left(\sum_{i=1}^{n+1}\Tilde{w}_i\cdot\delta_{S_i} \right)\\
        &\Leftrightarrow \mathbb{Q}_{1-\alpha} \left(\sum_{i=1}^{n}\Tilde{w}_i\cdot\delta_{S_i} +\Tilde{w}_{n+1}\cdot\delta_{+\infty}\right)
    \end{split}
\end{align}
Then define an \textit{unusual set} function $\mathcal{U}$:
\begin{equation}
    \mathcal{U}(S) = \left\{i\in[n+1]: S_i>\mathbb{Q}_{1-\alpha} \left(\sum_{i=1}^{n+1}\Tilde{w}_i\cdot\delta_{S_i} \right)\right\},
\end{equation}
which represents the indices $i$ where the deep nonconformity score values are too large. Then we know that noncoverage of $\bs Y_{n+1}$ implies the unusualness of point $k$:
\begin{equation}
    \bs Y_{n+1}\notin C_\theta(\bs X_{n+1})\rightarrow k\in \mathcal{U}(\bs S_\theta(\bs Z^k)).
\end{equation}

Thus, we have:
\begin{align}
\begin{split}
    \mathbb{P}(k\in \mathcal{U}(\bs S_\theta(\bs Z^k))) &= \sum_{i=1}^{n+1}\mathbb{P}\left(k=i, i\in\mathcal{U}(\bs S_\theta(\bs Z^i)\right)\\
    &=\sum_{i=1}^{n+1}\Tilde{w}_i\cdot \mathbb{P}(i\in\mathcal{U}(\bs S_\theta(\bs Z^i))\\
    &\leq \sum_{i=1}^{n+1}\Tilde{w}_i\left(\mathbb{P}(i\in\mathcal{U}(\bs S_\theta(\bs Z)) + \dtv{\bs S_\theta(\bs Z)}{\bs S_\theta(\bs Z^i)}\right)\\
    &= \mathbb{E}\left[\sum_{i\sim\mathcal{U}(\bs S_\theta(\bs Z^i))}\Tilde{w}_i\right]+\sum_{i=1}^{n}\Tilde{w}_i\dtv{\bs S_\theta(\bs Z)}{\bs S_\theta(\bs Z^i)}\\
    &\leq \alpha + \sum_{i=1}^{n}\Tilde{w}_i\dtv{\bs S_\theta(\bs Z)}{\bs S_\theta(\bs Z^i)}\quad\mbox{[by Lemma \ref{lemma:simplebound}]}\\
    \mbox{By complement, }\mathbb{P}\left(\bs Y_{n+1}\in C_\theta(\bs X_{n+1})\right) &\geq 1-\alpha-\sum_{i=1}^{n}\Tilde{w}_i\cdot \dtv{\bs S_\theta(\bs Z)}{\bs S_\theta(\bs Z^i)}
\end{split}
\end{align}
\endproof

\section{Upper Bounds of Coverage Gap}
\label{app:secB}
We provide two possible bounds for the coverage gap. The first bound comes from an example usage in \citet{Barber23aos-conformal} and the second bound is derived by assuming the outputs follow a beta distribution.
\subsection{Bounded Periodic Changes}
We are interested in finding an upper bound of the coverage gap in \cref{thm:coverage}. Specifically, we are trying to bound $\dtv{\bs S_\theta(\bs Z)}{\bs S_\theta(\bs Z^i)}$. One interesting case to analyze for video data is that we might have periodic large changes in the distribution rather than a gradual drift (i.e. there might 
 be a changepoint in the calibration set). 
 
 \begin{assumption}[$k$-step changepoints in dataset]
 \label{assump:changepoint}
 Suppose that the most recent
changepoint occurred $k$ steps ago, so that $\dtv{\bs S_\theta(\bs Z_{n+1})}{\bs S_\theta(\bs Z_i)}\rightarrow0$ for $i > n - k$. However, we might have an arbitrarily large difference from the new test point before the changepoint step: $\dtv{\bs S_\theta(\bs Z_{n+1})}{\bs S_\theta(\bs Z_i)}\rightarrow1$ for $i \leq n - k$.
 \end{assumption}
 
Now, we try to bound the coverage gap from \cref{thm:coverage}. We first design the weights amenable to our analysis. Recall the definition of weights:
\sdw*
Again, recall the miscoverage gap under bounded period changes:
\gapone*


\emph{Proof.}
\proof
We follow the sketch proof in \cite{Barber23aos-conformal} and provide detailed proof for interested readers. From Assumption \ref{assump:changepoint}, suppose the variation before the changepoint could be arbitrarily large:
\begin{align}
    \begin{split}
        \sum_{i=1}^{n}\Tilde{w}_i\dtv{\bs S_\theta(\bs Z)}{\bs S_\theta(\bs Z^i)}&\leq \sum_{i=1}^{n-k}\Tilde{w}_i\\
        &= \frac{\sum_{i=1}^{n-k}\rho^{n+1-\pi(i)}}{1+\sum_{i=1}^{n}\rho^{n+1-\pi(i)}}\\
        &\leq \rho^k\cdot\frac{1-\rho^{n-k}}{1-\rho^n}\\
        &\leq \rho^k
    \end{split}
\end{align}
\endproof
Intuitively, this tells us that the coverage gap will be small as long as $k$ is large - namely, as long as we have enough data after the changepoint.

\begin{remark}[Measuring $\rho$ and $k$]
    As explained in \citet{Barber23aos-conformal}, $\rho$ is a decay parameter less than 1 and the above miscoverage gap is small as long as $k$ is sufficiently large. To measure $k$, for each of the three test datasets in \cref{coverage-ablation}, we measure the average number of video sequences between the two closest sequence datapoints belonging to different subjects/activities.
\end{remark}

\subsection{Bounded Covariates}
Next, we are trying to bound $\dtv{\bs S_\theta(\bs Z)}{\bs S_\theta(\bs Z^i)}$ under distributional modeling. To find this bound, we make use of the \textbf{Hellinger distance}. The Hellinger distance $H^2(\bs P, \bs Q)$ between two probability measure $\bs P$ and $\bs Q$ on a measure space $\mathcal{X}$ with respect to an auxiliary measure $\lambda$ (e.g. joint) is defined as:
\[
H^2(\bs P, \bs Q) = \frac{1}{2}\int_\mathcal{X}\left(\sqrt{p(x)} - \sqrt{q(x)}\right)^2 \lambda(dx),
\]
where $\bs P(dx) = p(x) \lambda(dx)$ and $\bs Q(dx) = q(x) \lambda(dx)$. Succinctly, we can denote:
\[
H(\bs P, \bs Q) = \frac{1}{\sqrt{2}}||\sqrt{\bs P} - \sqrt{\bs Q}||_2
\]
It turns out that we can make use of the Hellinger distance to bound the total variation distance.
\begin{equation}
\label{lemma:hellinger-tv}
\dtv{\bs P}{\bs Q}\leq \sqrt{2}H(\bs P, \bs Q) 
\end{equation}

\emph{Proof.}
\proof
This is a fairly well-known results in statistics and we provide a detailed proof for the sake of completeness.
\begin{align}
\begin{split}
    \dtvv{\bs P}{\bs Q} &= \frac{1}{4}\left(\sum_i|p_i-q_i|^2\right)\\
    &= \frac{1}{4}\left(\sum_i(\sqrt{p_i}-\sqrt{q_i})(\sqrt{p_i}+\sqrt{q_i})\right)^2\\
    &\leq \frac{1}{4}\left(\sum_i(\sqrt{p_i}-\sqrt{q_i})^2\right)\left(\sum_i(\sqrt{p_i}+\sqrt{q_i})^2\right)\\
    &\leq \frac{1}{2}H^2(\bs P, \bs Q)\left(2+2\sum_i\sqrt{p_i}\sqrt{q_i}\right)\\
    &\leq H^2(\bs P, \bs Q)(2-H^2(\bs P, \bs Q))\leq 2H^2(\bs P, \bs Q)\\
   \text{Hence, } \dtv{\bs P}{\bs Q}&\leq \sqrt{2}H(\bs P, \bs Q)
\end{split}
\end{align}
\endproof
We can now focus on providing an upper-bound Hellinger distance instead. Since our conformity score is outputted by a sigmoid function via Monte Carlo Dropout during test time, one reasonable assumption is that the conformity score outputs follow a Beta distribution.
\begin{assumption}[Conformity scores follow a Beta distribution]
\label{asmp:boundedbeta}
\[
\bs S_\theta(\bs Z) \sim \beta(a_1, n-a_1), \quad \bs S_\theta(\bs Z^i) \sim \beta(a_2, n-a_2)
\]
where $a_1$ and $a_2$ are defined by the permuted dataset. Without loss of generality, we then assume:
    \[
    a_1-k\leq a_2\leq a_1+k
    \]
    That is, assume the difference in the Beta parameters is bounded.
\end{assumption}

The assumption makes intuitive sense in that it can be thought of as the proportion of calibration data points that are close to (\emph{conform to}) the new test point. Furthermore, the expected proportion change between the original and permuted dataset is bounded by $\frac{k}{n}$. This assumption makes sense in that we are essentially assuming that after swapping one pair of data points, the change in proportion of data that conform to the test data is bounded.

With the Beta distributions defined, recall the bound to prove:

\gaptwo*

\emph{Proof.}
\proof
First, we can express the Hellinger distance between two Beta-distributed measures in closed form:
\begin{equation}
\label{eq:def-hel-beta}
    H^2(\bs S_\theta(\bs Z), \bs S_\theta(\bs  Z^i)) = 1 - \frac{B\left(\frac{a_1+a_2}{2}, n-\frac{a_1+a_2}{2}\right)}{\sqrt{B(a_1, n-a_1)B(a_2, n-a_2)}},
\end{equation}
where $B(m, n)$ is the \textbf{beta function} defined as:
\begin{align}
\begin{split}
    B(m, n) &= \frac{\Gamma(m)\Gamma(n)}{\Gamma(m+n)}\\
    &=\frac{(m-1)!(n-1)!}{(m+n-1)!} = \frac{\frac{m+n}{mn}}{\binom{m+n}{n}}
\end{split}
\end{align}
We define a short-hand notation $a:=\frac{a_1+a_2}{2}$, and then by the definition of beta function, the numerator of Eq. \ref{eq:def-hel-beta} becomes:
\begin{equation}
    B\left(\frac{a_1+a_2}{2}, n-\frac{a_1+a_2}{2}\right) = \frac{(a-1)!(n-1-a)!}{(n-1)!}
\end{equation}
Similarly, in the denominator:
\begin{equation}
    B(a_1, n-a_1) = \frac{(a_1-1)!(n-1-a_1)!}{(n-1)!}, \quad B(a_2, n-a_2) = \frac{(a_2-1)!(n-1-a_2)!}{(n-1)!}
\end{equation}
Combining the expressions above and plugging them into Eq. \ref{eq:def-hel-beta}, we have:
\begin{align}
\begin{split}
\label{eq:hzz}
H^2(\bs S_\theta(\bs Z), \bs S_\theta(\bs  Z^i)) &= 1 - \frac{B\left(\frac{a_1+a_2}{2}, n-\frac{a_1+a_2}{2}\right)}{\sqrt{B(a_1, n-a_1)B(a_2, n-a_2)}}\\
&= 1-\frac{\frac{(a-1)!(n-1-a)!}{(n-1)!}}{\sqrt{\frac{(a_1-1)!(n-1-a_1)!}{(n-1)!}\cdot \frac{(a_2-1)!(n-1-a_2)!}{(n-1)!}}}\\
&= 1-\frac{(a-1)!(n-1-a)!}{\sqrt{(a_1-1)!(a_2-1)!(n-1-a_1)!(n-1-a_2)!}}
\end{split}
\end{align}


With Assumption \ref{asmp:boundedbeta} in place, we are able to bound the fraction in Eq. \ref{eq:hzz} as follows:
\begin{align}
\begin{split}
\left(\frac{(a-1)!(n-1-a)!}{\sqrt{(a_1-1)!(a_2-1)!(n-1-a_1)!(n-1-a_2)!}}\right)^2 &= \frac{\Gamma(a)\Gamma(a)\Gamma(n-a)\Gamma(n-a)}{\Gamma(a_1)\Gamma(a_2)\Gamma(n-a_1)\Gamma(n-a_2)}\\
\mbox{[by bounded difference between $a_1$ and $a_2$]}\quad&\geq  \frac{\Gamma(a_1+\frac{k}{2})\Gamma(a_1+\frac{k}{2})\Gamma(n-a_1-\frac{k}{2})\Gamma(n-a_1-\frac{k}{2})}{\Gamma(a_1)\Gamma(a_1+k)\Gamma(n-a_1)\Gamma(n-a_1-k)}\\
\mbox{[by definition of $\Gamma$ function]}\quad&\geq a_1^\frac{k}{2}\cdot(a_1+k)^{-\frac{k}{2}}\cdot (n-a_1)^{-\frac{k}{2}}\cdot(n-a_1-k)^{\frac{k}{2}}\\
\mbox{[rearranging the terms]}\quad&\geq \left(\frac{a_1\cdot(n-a_1-k)}{(a_1+k)\cdot(n-a_1)}\right)^\frac{k}{2}\\
\mbox{[expanding the terms]}\quad&\geq\left(\frac{a_1n-a_1^2-a_1k}{a_1n-a_1^2-a_1k+kn}\right)^\frac{k}{2}\\
&\geq \left(1-\frac{kn}{{a_1n-a_1^2-a_1k+kn}}\right)^\frac{k}{2}\\
\mbox{[denominator extremum at $a_1 = \frac{n-k}{2}$]}\quad&\geq \left(1 - \frac{kn}{\frac{(n-k)^2}{4}+kn}\right)^\frac{k}{2}\\
&\geq\left(\frac{(n-k)^2}{{(n-k)^2}+4kn}\right)^\frac{k}{2}\\
&\geq\left(\frac{n-k}{n+k}\right)^{k}\\
\mbox{Hence, }\frac{(a-1)!(n-1-a)!}{\sqrt{(a_1-1)!(a_2-1)!(n-1-a_1)!(n-1-a_2)!}}&\geq \left(1-\frac{2k}{{n+k}}\right)^\frac{k}{2}
\end{split}
\end{align}


Putting it all together, we know that
\[
1-H^2(\bs S_\theta(\bs Z), \bs S_\theta(\bs  Z^i))\geq \left(1-\frac{2k}{{n+k}}\right)^\frac{k}{2}
\]
and equivalently, 
\[
H^2(\bs S_\theta(\bs Z), \bs S_\theta(\bs  Z^i))\leq 1- \left(1-\frac{2k}{{n+k}}\right)^\frac{k}{2}
\]
Hence, we have the final upper bound on the coverage gap via Lemma \ref{lemma:hellinger-tv}:
\begin{equation}
    \dtv{\bs S_\theta(\bs Z)}{\bs S_\theta(\bs Z^i)}\leq \sqrt{2- 2\left(1-\frac{2k}{{n+k}}\right)^\frac{k}{2}}
\end{equation}

Now, we analyze the behavior of this bound. We rewrite the upper bound as follows:
\[
\sqrt{2}\cdot\sqrt{1- \left(1-\frac{2k}{{n+k}}\right)^\frac{k}{2}}
\]

We first define $p:=\frac{k}{n}$. Then we have:
\[
1-\frac{2k}{{n+k}} = \frac{n-k}{n+k} = \frac{1-p}{1+p}
\]
By Taylor series, we have the following series expansion:
\begin{align}
    \begin{split}
        \ln\left(\frac{1-p}{1+p}\right) = \ln(1-p) - \ln(1+p) &= -p-\frac{p^2}{2} - \frac{p^3}{3}+\cdots - \left(p-\frac{p^2}{2} + \frac{p^3}{3}+\cdots\right)\\
        &= -2p - \frac{2}{3}p^3+\cdots
    \end{split}
\end{align}
Then we have the following behavior:
\begin{align}
    \begin{split}
    1-\frac{2k}{{n+k}} = \left(\frac{1-p}{1+p}\right)^\frac{k}{2} &= \exp\left(\frac{k}{2}\ln\left(\frac{1-p}{1+p}\right)\right)\\
    &= \exp\left(\frac{k}{2}\left(-2p - \frac{2}{3}p^3+\cdots\right)\right)\\
    &\sim \exp\left(-kp-\frac{k}{3}kp^3\right)\\
    &\sim\exp\left(-\frac{k^2}{n}\right)
    \end{split}
\end{align}
Thus, we have:
\begin{equation}
    \dtv{\bs S_\theta(\bs Z)}{\bs S_\theta(\bs Z^i)}\leq \sqrt{2- 2\left(1-\frac{2k}{{n+k}}\right)^\frac{k}{2}} \sim\mathcal{O}\left(\sqrt{1-\exp\left(-\frac{k^2}{n}\right)}\right)
\end{equation}
\endproof
This behavior indicates that the bounds gets weaker exponentially with larger $k$.
\begin{remark}[Measuring $k$]
     The parameter $k$ in the bound can be measured empirically for each video calibration dataset. To measure $k$, for each of the three test datasets in \cref{coverage-ablation}, we measure the average changes of labels of subjects/activities after swapping the $i$-th data point with the last one. This value is usually low (in most cases $\leq 2$).
\end{remark}

\section{Training and Testing Datasets Details}
\label{traintestdataset}
We use the standard 3D human shape-pose datasets: 3DPW \cite{Von18eccv-3dpw, pan2022tax, pan2023tax}, Human3.6M \cite{Ionescu13tpami-human3, zhang2016health, zhang2020dex}, MPII-3DHP \cite{Mehta173dv-monocular, zhang2021robots, avigal20206}, Penn Action \cite{Zhang13iccv-penn, avigal2021avplug}, PoseTrack \cite{Andriluka18cvpr-posetrack, elmquist2022art, sim2019personalization, shen2024diffclip}, and InstaVariety \cite{Kanazawa19cvpr-learning, yao2023apla, zhang2023flowbot++, eisner2022flowbot3d} where the preprocessed data is provided by~\cite{Shen23cvpr-glot, devgon2020orienting, lim2021planar, lim2022real2sim2real}, \cite{Choi21cvpr-tcmr, wang2024self, sundaresan2024learning}, and \cite{Kocabas20cvpr-vibe}, and evaluated on 3DPW, Human3.6M, MPII-3DHP. Note that our training dataset is about 2.5\% smaller than previous works because we hold out a small portion ($\sim$ 1500 datapoints) for calibration.

\section{Details of Choice of Conformity Score Function}
\label{choice}
Note that all three mentioned conformity score functions were trained end-to-end with the training time ensemble augmentation setting since learning the score function after the human reconstructor is trained does not improve the performance. For the score function augmented with inefficiency loss, we are essentially controlling the size of the conformal prediction set during training \cite{Stutz21iclr-learning}. For the classifier-style conformity score function, the training objective is to classify if the mean 3D keypoints L2 loss is within 40mm from the groundtruth using BCE loss and we use the logit as the conformity score. It is worth noting that all three variants result in better performance, demonstrating the importance of training time ensemble augmentation. %
Note that Ineff. needs more proposals during training than others and converges more slowly.
\section{Limitations}
\label{app:limitation}
While CUPS performs well on various benchmarks, we acknowledge that it does have some limitations. First, many samples need to be proposed during training
to improve the learned nonconformity score, which consumes a lot more GPU memory (30\% more going from 10 proposals to 20) and slows down the training process. Second, the method does not utilize human joint-level information, which could potentially improve the performance.

%% file: main.bbl
\begin{thebibliography}{63}
\providecommand{\natexlab}[1]{#1}
\providecommand{\url}[1]{\texttt{#1}}
\expandafter\ifx\csname urlstyle\endcsname\relax
  \providecommand{\doi}[1]{doi: #1}\else
  \providecommand{\doi}{doi: \begingroup \urlstyle{rm}\Url}\fi

\bibitem[Andriluka et~al.(2018)Andriluka, Iqbal, Insafutdinov, Pishchulin, Milan, Gall, and Schiele]{Andriluka18cvpr-posetrack}
Andriluka, M., Iqbal, U., Insafutdinov, E., Pishchulin, L., Milan, A., Gall, J., and Schiele, B.
\newblock Posetrack: A benchmark for human pose estimation and tracking.
\newblock In \emph{IEEE Conf. on Computer Vision and Pattern Recognition (CVPR)}, pp.\  5167--5176, 2018.

\bibitem[Angelopoulos et~al.(2024)Angelopoulos, Candes, and Tibshirani]{Angelopoulos24neurips-conformal}
Angelopoulos, A., Candes, E., and Tibshirani, R.~J.
\newblock Conformal pid control for time series prediction.
\newblock \emph{Conf. on Neural Information Processing Systems (NeurIPS)}, 36, 2024.

\bibitem[Angelopoulos \& Bates(2021)Angelopoulos and Bates]{Angelopoulos21gentle}
Angelopoulos, A.~N. and Bates, S.
\newblock A gentle introduction to conformal prediction and distribution-free uncertainty quantification.
\newblock \emph{arXiv preprint arXiv:2107.07511}, 2021.

\bibitem[Angelopoulos et~al.(2022)Angelopoulos, Kohli, Bates, Jordan, Malik, Alshaabi, Upadhyayula, and Romano]{Angelopoulos22icml-image}
Angelopoulos, A.~N., Kohli, A.~P., Bates, S., Jordan, M., Malik, J., Alshaabi, T., Upadhyayula, S., and Romano, Y.
\newblock Image-to-image regression with distribution-free uncertainty quantification and applications in imaging.
\newblock In \emph{Intl. Conf. on Machine Learning (ICML)}, pp.\  717--730. PMLR, 2022.

\bibitem[Avigal et~al.(2020)Avigal, Paradis, and Zhang]{avigal20206}
Avigal, Y., Paradis, S., and Zhang, H.
\newblock 6-dof grasp planning using fast 3d reconstruction and grasp quality cnn.
\newblock \emph{arXiv preprint arXiv:2009.08618}, 2020.

\bibitem[Avigal et~al.(2021)Avigal, Satish, Tam, Huang, Zhang, Danielczuk, Ichnowski, and Goldberg]{avigal2021avplug}
Avigal, Y., Satish, V., Tam, Z., Huang, H., Zhang, H., Danielczuk, M., Ichnowski, J., and Goldberg, K.
\newblock Avplug: Approach vector planning for unicontact grasping amid clutter.
\newblock In \emph{2021 IEEE 17th International Conference on Automation Science and Engineering (CASE)}, pp.\  1140--1147. IEEE, 2021.

\bibitem[Barber et~al.(2023)Barber, Candes, Ramdas, and Tibshirani]{Barber23aos-conformal}
Barber, R.~F., Candes, E.~J., Ramdas, A., and Tibshirani, R.~J.
\newblock Conformal prediction beyond exchangeability.
\newblock \emph{The Annals of Statistics}, 51\penalty0 (2):\penalty0 816--845, 2023.

\bibitem[Biggs et~al.(2020)Biggs, Novotny, Ehrhardt, Joo, Graham, and Vedaldi]{Biggs20neurips-3dmb}
Biggs, B., Novotny, D., Ehrhardt, S., Joo, H., Graham, B., and Vedaldi, A.
\newblock 3d multi-bodies: Fitting sets of plausible 3d human models to ambiguous image data.
\newblock \emph{Conf. on Neural Information Processing Systems (NeurIPS)}, 33:\penalty0 20496--20507, 2020.

\bibitem[Choi et~al.(2021)Choi, Moon, Chang, and Lee]{Choi21cvpr-tcmr}
Choi, H., Moon, G., Chang, J.~Y., and Lee, K.~M.
\newblock Beyond static features for temporally consistent 3d human pose and shape from a video.
\newblock In \emph{IEEE Conf. on Computer Vision and Pattern Recognition (CVPR)}, pp.\  1964--1973, 2021.

\bibitem[Ci et~al.(2019)Ci, Wang, Ma, and Wang]{Ci19iccv-optimizing}
Ci, H., Wang, C., Ma, X., and Wang, Y.
\newblock Optimizing network structure for 3d human pose estimation.
\newblock In \emph{Intl. Conf. on Computer Vision (ICCV)}, pp.\  2262--2271, 2019.

\bibitem[Devgon et~al.(2020)Devgon, Ichnowski, Balakrishna, Zhang, and Goldberg]{devgon2020orienting}
Devgon, S., Ichnowski, J., Balakrishna, A., Zhang, H., and Goldberg, K.
\newblock Orienting novel 3d objects using self-supervised learning of rotation transforms.
\newblock In \emph{2020 IEEE 16th International Conference on Automation Science and Engineering (CASE)}, pp.\  1453--1460. IEEE, 2020.

\bibitem[Dwivedi et~al.(2024)Dwivedi, Schmid, Yi, Black, and Tzionas]{Dwivedi243dv-poco}
Dwivedi, S.~K., Schmid, C., Yi, H., Black, M.~J., and Tzionas, D.
\newblock Poco: 3d pose and shape estimation with confidence.
\newblock In \emph{2024 International Conference on 3D Vision (3DV)}, pp.\  85--95. IEEE, 2024.

\bibitem[Eisner et~al.(2022)Eisner, Zhang, and Held]{eisner2022flowbot3d}
Eisner, B., Zhang, H., and Held, D.
\newblock Flowbot3d: Learning 3d articulation flow to manipulate articulated objects.
\newblock \emph{arXiv preprint arXiv:2205.04382}, 2022.

\bibitem[Elmquist et~al.(2022)Elmquist, Young, Hansen, Ashokkumar, Caldararu, Dashora, Mahajan, Zhang, Fang, Shen, et~al.]{elmquist2022art}
Elmquist, A., Young, A., Hansen, T., Ashokkumar, S., Caldararu, S., Dashora, A., Mahajan, I., Zhang, H., Fang, L., Shen, H., et~al.
\newblock Art/atk: A research platform for assessing and mitigating the sim-to-real gap in robotics and autonomous vehicle engineering.
\newblock \emph{arXiv preprint arXiv:2211.04886}, 2022.

\bibitem[Gal \& Ghahramani(2016)Gal and Ghahramani]{Gal16icml-mcdropout}
Gal, Y. and Ghahramani, Z.
\newblock Dropout as a bayesian approximation: Representing model uncertainty in deep learning.
\newblock In \emph{Intl. Conf. on Machine Learning (ICML)}, pp.\  1050--1059. PMLR, 2016.

\bibitem[Goodfellow et~al.(2020)Goodfellow, Pouget-Abadie, Mirza, Xu, Warde-Farley, Ozair, Courville, and Bengio]{Goodfellow20acm-generative}
Goodfellow, I., Pouget-Abadie, J., Mirza, M., Xu, B., Warde-Farley, D., Ozair, S., Courville, A., and Bengio, Y.
\newblock Generative adversarial networks.
\newblock \emph{Communications of the ACM}, 63\penalty0 (11):\penalty0 139--144, 2020.

\bibitem[Harrison(2012)]{Harrison12conservative}
Harrison, M.~T.
\newblock Conservative hypothesis tests and confidence intervals using importance sampling.
\newblock \emph{Biometrika}, 99\penalty0 (1):\penalty0 57--69, 2012.

\bibitem[Ho et~al.(2022)Ho, Saharia, Chan, Fleet, Norouzi, and Salimans]{Ho22jmlr-cascaded}
Ho, J., Saharia, C., Chan, W., Fleet, D.~J., Norouzi, M., and Salimans, T.
\newblock Cascaded diffusion models for high fidelity image generation.
\newblock \emph{J. of Machine Learning Research}, 23\penalty0 (47):\penalty0 1--33, 2022.

\bibitem[Holmquist \& Wandt(2023)Holmquist and Wandt]{Holmquist23iccv-diffpose}
Holmquist, K. and Wandt, B.
\newblock Diffpose: Multi-hypothesis human pose estimation using diffusion models.
\newblock In \emph{Intl. Conf. on Computer Vision (ICCV)}, pp.\  15977--15987, 2023.

\bibitem[Ionescu et~al.(2013)Ionescu, Papava, Olaru, and Sminchisescu]{Ionescu13tpami-human3}
Ionescu, C., Papava, D., Olaru, V., and Sminchisescu, C.
\newblock Human3. 6m: Large scale datasets and predictive methods for 3d human sensing in natural environments.
\newblock \emph{{IEEE} Trans. Pattern Anal. Machine Intell.}, 36\penalty0 (7):\penalty0 1325--1339, 2013.

\bibitem[Kanazawa et~al.(2018)Kanazawa, Black, Jacobs, and Malik]{Kanazawa18cvpr-smplify}
Kanazawa, A., Black, M.~J., Jacobs, D.~W., and Malik, J.
\newblock End-to-end recovery of human shape and pose.
\newblock In \emph{IEEE Conf. on Computer Vision and Pattern Recognition (CVPR)}, pp.\  7122--7131, 2018.

\bibitem[Kanazawa et~al.(2019)Kanazawa, Zhang, Felsen, and Malik]{Kanazawa19cvpr-learning}
Kanazawa, A., Zhang, J.~Y., Felsen, P., and Malik, J.
\newblock Learning 3d human dynamics from video.
\newblock In \emph{IEEE Conf. on Computer Vision and Pattern Recognition (CVPR)}, pp.\  5614--5623, 2019.

\bibitem[Kocabas et~al.(2020)Kocabas, Athanasiou, and Black]{Kocabas20cvpr-vibe}
Kocabas, M., Athanasiou, N., and Black, M.~J.
\newblock Vibe: Video inference for human body pose and shape estimation.
\newblock In \emph{IEEE Conf. on Computer Vision and Pattern Recognition (CVPR)}, pp.\  5253--5263, 2020.

\bibitem[Kolotouros et~al.(2019)Kolotouros, Pavlakos, Black, and Daniilidis]{Kolotouros19iccv-spin}
Kolotouros, N., Pavlakos, G., Black, M.~J., and Daniilidis, K.
\newblock Learning to reconstruct 3d human pose and shape via model-fitting in the loop.
\newblock In \emph{Intl. Conf. on Computer Vision (ICCV)}, pp.\  2252--2261, 2019.

\bibitem[Li \& Lee(2019)Li and Lee]{Li19cvpr-generating}
Li, C. and Lee, G.~H.
\newblock Generating multiple hypotheses for 3d human pose estimation with mixture density network.
\newblock In \emph{IEEE Conf. on Computer Vision and Pattern Recognition (CVPR)}, pp.\  9887--9895, 2019.

\bibitem[Lim et~al.(2021)Lim, Huang, Chen, Wang, Ichnowski, Seita, Laskey, and Goldberg]{lim2021planar}
Lim, V., Huang, H., Chen, L.~Y., Wang, J., Ichnowski, J., Seita, D., Laskey, M., and Goldberg, K.
\newblock Planar robot casting with real2sim2real self-supervised learning.
\newblock \emph{arXiv preprint arXiv:2111.04814}, 2021.

\bibitem[Lim et~al.(2022)Lim, Huang, Chen, Wang, Ichnowski, Seita, Laskey, and Goldberg]{lim2022real2sim2real}
Lim, V., Huang, H., Chen, L.~Y., Wang, J., Ichnowski, J., Seita, D., Laskey, M., and Goldberg, K.
\newblock Real2sim2real: Self-supervised learning of physical single-step dynamic actions for planar robot casting.
\newblock In \emph{2022 International Conference on Robotics and Automation (ICRA)}, pp.\  8282--8289. IEEE, 2022.

\bibitem[Loper et~al.(2023)Loper, Mahmood, Romero, Pons-Moll, and Black]{Loper23sg-smpl}
Loper, M., Mahmood, N., Romero, J., Pons-Moll, G., and Black, M.~J.
\newblock Smpl: A skinned multi-person linear model.
\newblock In \emph{Seminal Graphics Papers: Pushing the Boundaries, Volume 2}, pp.\  851--866. 2023.

\bibitem[Luo et~al.(2020)Luo, Golestaneh, and Kitani]{Luo20accv-meva}
Luo, Z., Golestaneh, S.~A., and Kitani, K.~M.
\newblock 3d human motion estimation via motion compression and refinement.
\newblock In \emph{Asian Conf. on Computer Vision (ACCV)}, 2020.

\bibitem[Ma et~al.(2021)Ma, Su, Wang, Ci, and Wang]{Ma21cvpr-context}
Ma, X., Su, J., Wang, C., Ci, H., and Wang, Y.
\newblock Context modeling in 3d human pose estimation: A unified perspective.
\newblock In \emph{IEEE Conf. on Computer Vision and Pattern Recognition (CVPR)}, pp.\  6238--6247, 2021.

\bibitem[Ma et~al.(2022)Ma, Rahmani, Fan, Yang, Chen, and Liu]{Ma22aaai-remote}
Ma, X., Rahmani, H., Fan, Z., Yang, B., Chen, J., and Liu, J.
\newblock Remote: Reinforced motion transformation network for semi-supervised 2d pose estimation in videos.
\newblock In \emph{Nat. Conf. on Artificial Intelligence (AAAI)}, volume~36, pp.\  1944--1952, 2022.

\bibitem[Mehta et~al.(2017)Mehta, Rhodin, Casas, Fua, Sotnychenko, Xu, and Theobalt]{Mehta173dv-monocular}
Mehta, D., Rhodin, H., Casas, D., Fua, P., Sotnychenko, O., Xu, W., and Theobalt, C.
\newblock Monocular 3d human pose estimation in the wild using improved cnn supervision.
\newblock In \emph{2017 international conference on 3D vision (3DV)}, pp.\  506--516. IEEE, 2017.

\bibitem[Pan et~al.(2022)Pan, Okorn, Zhang, Eisner, and Held]{pan2022tax}
Pan, C., Okorn, B., Zhang, H., Eisner, B., and Held, D.
\newblock Tax-pose: Task-specific cross-pose estimation for robot manipulation.
\newblock \emph{arXiv preprint arXiv:2211.09325}, 2022.

\bibitem[Pan et~al.(2023)Pan, Okorn, Zhang, Eisner, and Held]{pan2023tax}
Pan, C., Okorn, B., Zhang, H., Eisner, B., and Held, D.
\newblock Tax-pose: Task-specific cross-pose estimation for robot manipulation.
\newblock In \emph{Conference on Robot Learning}, pp.\  1783--1792. PMLR, 2023.

\bibitem[Pavlakos et~al.(2017)Pavlakos, Zhou, Derpanis, and Daniilidis]{Pavlakos17cvpr-coarse}
Pavlakos, G., Zhou, X., Derpanis, K.~G., and Daniilidis, K.
\newblock Coarse-to-fine volumetric prediction for single-image 3d human pose.
\newblock In \emph{IEEE Conf. on Computer Vision and Pattern Recognition (CVPR)}, pp.\  7025--7034, 2017.

\bibitem[Rempe et~al.(2021)Rempe, Birdal, Hertzmann, Yang, Sridhar, and Guibas]{Rempe21cvpr-humor}
Rempe, D., Birdal, T., Hertzmann, A., Yang, J., Sridhar, S., and Guibas, L.~J.
\newblock Humor: 3d human motion model for robust pose estimation.
\newblock In \emph{IEEE Conf. on Computer Vision and Pattern Recognition (CVPR)}, pp.\  11488--11499, 2021.

\bibitem[Shafer \& Vovk(2008)Shafer and Vovk]{Shafer08jmlr-tutorial}
Shafer, G. and Vovk, V.
\newblock A tutorial on conformal prediction.
\newblock \emph{J. of Machine Learning Research}, 9\penalty0 (3), 2008.

\bibitem[Shan et~al.(2023)Shan, Liu, Zhang, Wang, Han, Wang, Ma, and Gao]{Shan23cvpr-diffusion}
Shan, W., Liu, Z., Zhang, X., Wang, Z., Han, K., Wang, S., Ma, S., and Gao, W.
\newblock Diffusion-based 3d human pose estimation with multi-hypothesis aggregation.
\newblock In \emph{IEEE Conf. on Computer Vision and Pattern Recognition (CVPR)}, pp.\  14761--14771, 2023.

\bibitem[Sharma et~al.(2019)Sharma, Varigonda, Bindal, Sharma, and Jain]{Sharma19iccv-monocular}
Sharma, S., Varigonda, P.~T., Bindal, P., Sharma, A., and Jain, A.
\newblock Monocular 3d human pose estimation by generation and ordinal ranking.
\newblock In \emph{Intl. Conf. on Computer Vision (ICCV)}, pp.\  2325--2334, 2019.

\bibitem[Shen et~al.(2024)Shen, Zhu, Fan, Zhang, and Wu]{shen2024diffclip}
Shen, S., Zhu, Z., Fan, L., Zhang, H., and Wu, X.
\newblock Diffclip: Leveraging stable diffusion for language grounded 3d classification.
\newblock In \emph{Proceedings of the IEEE/CVF Winter Conference on Applications of Computer Vision}, pp.\  3596--3605, 2024.

\bibitem[Shen et~al.(2023)Shen, Yang, Wang, Ma, Zhou, and Yang]{Shen23cvpr-glot}
Shen, X., Yang, Z., Wang, X., Ma, J., Zhou, C., and Yang, Y.
\newblock Global-to-local modeling for video-based 3d human pose and shape estimation.
\newblock In \emph{IEEE Conf. on Computer Vision and Pattern Recognition (CVPR)}, pp.\  8887--8896, 2023.

\bibitem[Sim et~al.(2019)Sim, Beaufays, Benard, Guliani, Kabel, Khare, Lucassen, Zadrazil, Zhang, Johnson, et~al.]{sim2019personalization}
Sim, K.~C., Beaufays, F., Benard, A., Guliani, D., Kabel, A., Khare, N., Lucassen, T., Zadrazil, P., Zhang, H., Johnson, L., et~al.
\newblock Personalization of end-to-end speech recognition on mobile devices for named entities.
\newblock In \emph{2019 IEEE Automatic Speech Recognition and Understanding Workshop (ASRU)}, pp.\  23--30. IEEE, 2019.

\bibitem[Stutz et~al.(2021)Stutz, Cemgil, Doucet, et~al.]{Stutz21iclr-learning}
Stutz, D., Cemgil, A.~T., Doucet, A., et~al.
\newblock Learning optimal conformal classifiers.
\newblock \emph{Intl. Conf. on Learning Representations (ICLR)}, 2021.

\bibitem[Sun et~al.(2024)Sun, Jiang, Qiu, Nobel, Kochenderfer, and Schwager]{Sun24neurips-conformal}
Sun, J., Jiang, Y., Qiu, J., Nobel, P., Kochenderfer, M.~J., and Schwager, M.
\newblock Conformal prediction for uncertainty-aware planning with diffusion dynamics model.
\newblock \emph{Conf. on Neural Information Processing Systems (NeurIPS)}, 36, 2024.

\bibitem[Sun et~al.(2018)Sun, Xiao, Wei, Liang, and Wei]{Sun18eccv-integral}
Sun, X., Xiao, B., Wei, F., Liang, S., and Wei, Y.
\newblock Integral human pose regression.
\newblock In \emph{European Conf. on Computer Vision (ECCV)}, pp.\  529--545, 2018.

\bibitem[Sundaresan et~al.(2024)Sundaresan, Ganapathi, Zhang, and Devgon]{sundaresan2024learning}
Sundaresan, P., Ganapathi, A., Zhang, H., and Devgon, S.
\newblock Learning correspondence for deformable objects.
\newblock \emph{arXiv preprint arXiv:2405.08996}, 2024.

\bibitem[Von~Marcard et~al.(2018)Von~Marcard, Henschel, Black, Rosenhahn, and Pons-Moll]{Von18eccv-3dpw}
Von~Marcard, T., Henschel, R., Black, M.~J., Rosenhahn, B., and Pons-Moll, G.
\newblock Recovering accurate 3d human pose in the wild using imus and a moving camera.
\newblock In \emph{Proceedings of the European conference on computer vision (ECCV)}, pp.\  601--617, 2018.

\bibitem[Wang et~al.(2024)Wang, Huang, Lim, Zhang, Ichnowski, Seita, Chen, and Goldberg]{wang2024self}
Wang, J., Huang, H., Lim, V., Zhang, H., Ichnowski, J., Seita, D., Chen, Y., and Goldberg, K.
\newblock Self-supervised learning of dynamic planar manipulation of free-end cables.
\newblock \emph{arXiv preprint arXiv:2405.09581}, 2024.

\bibitem[Wehrbein et~al.(2021)Wehrbein, Rudolph, Rosenhahn, and Wandt]{Wehrbein21iccv-probabilistic}
Wehrbein, T., Rudolph, M., Rosenhahn, B., and Wandt, B.
\newblock Probabilistic monocular 3d human pose estimation with normalizing flows.
\newblock In \emph{Intl. Conf. on Computer Vision (ICCV)}, pp.\  11199--11208, 2021.

\bibitem[Wei et~al.(2022)Wei, Lin, Liu, and Liao]{Wei22cvpr-mps}
Wei, W.-L., Lin, J.-C., Liu, T.-L., and Liao, H.-Y.~M.
\newblock Capturing humans in motion: Temporal-attentive 3d human pose and shape estimation from monocular video.
\newblock In \emph{IEEE Conf. on Computer Vision and Pattern Recognition (CVPR)}, pp.\  13211--13220, 2022.

\bibitem[Xu \& Takano(2021)Xu and Takano]{Xu21cvpr-graph}
Xu, T. and Takano, W.
\newblock Graph stacked hourglass networks for 3d human pose estimation.
\newblock In \emph{IEEE Conf. on Computer Vision and Pattern Recognition (CVPR)}, pp.\  16105--16114, 2021.

\bibitem[Yang \& Pavone(2023)Yang and Pavone]{Yang23cvpr-object}
Yang, H. and Pavone, M.
\newblock Object pose estimation with statistical guarantees: Conformal keypoint detection and geometric uncertainty propagation.
\newblock In \emph{IEEE Conf. on Computer Vision and Pattern Recognition (CVPR)}, pp.\  8947--8958, 2023.

\bibitem[Yao et~al.(2023)Yao, Deng, Cao, Zhang, and Deng]{yao2023apla}
Yao, Y., Deng, S., Cao, Z., Zhang, H., and Deng, L.-J.
\newblock Apla: Additional perturbation for latent noise with adversarial training enables consistency.
\newblock \emph{arXiv preprint arXiv:2308.12605}, 2023.

\bibitem[Zhan et~al.(2022)Zhan, Li, Weng, and Choi]{Zhan22cvpr-ray3d}
Zhan, Y., Li, F., Weng, R., and Choi, W.
\newblock Ray3d: ray-based 3d human pose estimation for monocular absolute 3d localization.
\newblock In \emph{IEEE Conf. on Computer Vision and Pattern Recognition (CVPR)}, pp.\  13116--13125, 2022.

\bibitem[Zhang(2016)]{zhang2016health}
Zhang, H.
\newblock Health diagnosis based on analysis of data captured by wearable technology devices.
\newblock \emph{International Journal of Advanced Science and Technology}, 95:\penalty0 89--96, 2016.

\bibitem[Zhang \& Carlone(2024)Zhang and Carlone]{Zhang24arxiv-CHAMP}
Zhang, H. and Carlone, L.
\newblock {CHAMP}: Conformalized {3D} human multi-hypothesis pose estimators.
\newblock \emph{arXiv preprint: 2407.06141}, 2024.

\bibitem[Zhang et~al.(2020)Zhang, Ichnowski, Avigal, Gonzales, Stoica, and Goldberg]{zhang2020dex}
Zhang, H., Ichnowski, J., Avigal, Y., Gonzales, J., Stoica, I., and Goldberg, K.
\newblock Dex-net ar: Distributed deep grasp planning using a commodity cellphone and augmented reality app.
\newblock In \emph{2020 IEEE International Conference on Robotics and Automation (ICRA)}, pp.\  552--558. IEEE, 2020.

\bibitem[Zhang et~al.(2021)Zhang, Ichnowski, Seita, Wang, Huang, and Goldberg]{zhang2021robots}
Zhang, H., Ichnowski, J., Seita, D., Wang, J., Huang, H., and Goldberg, K.
\newblock Robots of the lost arc: Self-supervised learning to dynamically manipulate fixed-endpoint cables.
\newblock In \emph{2021 IEEE International Conference on Robotics and Automation (ICRA)}, pp.\  4560--4567. IEEE, 2021.

\bibitem[Zhang et~al.(2023{\natexlab{a}})Zhang, Eisner, and Held]{zhang2023flowbot++}
Zhang, H., Eisner, B., and Held, D.
\newblock Flowbot++: Learning generalized articulated objects manipulation via articulation projection.
\newblock \emph{arXiv preprint arXiv:2306.12893}, 2023{\natexlab{a}}.

\bibitem[Zhang et~al.(2022)Zhang, Tu, Yang, Chen, and Yuan]{Zhang22cvpr-mixste}
Zhang, J., Tu, Z., Yang, J., Chen, Y., and Yuan, J.
\newblock Mixste: Seq2seq mixed spatio-temporal encoder for 3d human pose estimation in video.
\newblock In \emph{IEEE Conf. on Computer Vision and Pattern Recognition (CVPR)}, pp.\  13232--13242, 2022.

\bibitem[Zhang et~al.(2024)Zhang, Bhatnagar, Xu, Winkler, Kadlecek, Tang, and Bogo]{Zhang24cvpr-rohm}
Zhang, S., Bhatnagar, B.~L., Xu, Y., Winkler, A., Kadlecek, P., Tang, S., and Bogo, F.
\newblock Rohm: Robust human motion reconstruction via diffusion.
\newblock In \emph{IEEE Conf. on Computer Vision and Pattern Recognition (CVPR)}, pp.\  14606--14617, 2024.

\bibitem[Zhang et~al.(2013)Zhang, Zhu, and Derpanis]{Zhang13iccv-penn}
Zhang, W., Zhu, M., and Derpanis, K.~G.
\newblock From actemes to action: A strongly-supervised representation for detailed action understanding.
\newblock In \emph{Intl. Conf. on Computer Vision (ICCV)}, pp.\  2248--2255, 2013.

\bibitem[Zhang et~al.(2023{\natexlab{b}})Zhang, Wang, Kephart, and Ji]{Zhang23cvpr-body}
Zhang, Y., Wang, H., Kephart, J.~O., and Ji, Q.
\newblock Body knowledge and uncertainty modeling for monocular 3d human body reconstruction.
\newblock In \emph{IEEE Conf. on Computer Vision and Pattern Recognition (CVPR)}, pp.\  9020--9032, 2023{\natexlab{b}}.

\end{thebibliography}
